\documentclass[letterpaper, 10 pt, conference]{ieeeconf} 

\IEEEoverridecommandlockouts 

\usepackage{graphics} 
\usepackage{epsfig} 
\usepackage{mathptmx} 
\usepackage{times} 
\usepackage{amsmath} 
\usepackage{amssymb} 
\usepackage[skip=0.3ex]{subcaption}
\usepackage{placeins}
\usepackage[textwidth=1.9cm]{todonotes}
\title{\LARGE \bf
Smoother Position-Drift Compensation for Time Domain Passivity Approach based Teleoperation
}

\author{Andre Coelho, Harsimran Singh, Tin Muskardin, Ribin Balachandran, and Konstantin Kondak
\thanks{The authors are with the Institute of Robotics and Mechatronics of the
German Aerospace Center (DLR), Oberpfaffenhofen, Germany
{\tt\small andre.coelho@dlr.de}}%
}
\begin{document}

\maketitle
\thispagestyle{empty}
\pagestyle{empty}

\begin{abstract}
Despite being one of the most robust methods in bilateral teleoperation, Time Domain Passivity Approach (TDPA) presents the drawback of accumulating position drift between master and slave devices. The lack of position synchronization poses an obstacle to the performance of teleoperation and may prevent the successful accomplishment of such tasks. Several techniques have been developed in order to solve the position-drift problem in TDPA-based teleoperation. However, they either present poor transparency by over-conservatively constraining force feedback or add high impulse-like force signals that can be harmful to the hardware and to the human operator. We propose a new approach to compensate position drift in TDPA-based teleoperation in a smoother way, which keeps the forces within the normal range of the teleoperation task while preserving the level of transparency and the robust stability of energy-based TDPA. We also add a way of tuning the compensator to behave in accordance with the task being performed, whether it requires faster or smoother compensation. The feasibility and performance of the method were experimentally validated. Good position tracking and regular-amplitude forces are demonstrated with up to 500 ms round-trip constant and variable delays for hard-wall contacts. 
\end{abstract}

\section{INTRODUCTION}

Stability of teleoperation systems has been a concern since the beginning of their application. In a feedback loop including two stable systems, the stability of the individual systems does not guarantee overall stability. Adding to that, the connection between master and slave devices in teleoperation is usually accomplished by a communication channel with time delay. This often compromises the overall stability. To solve that, different techniques based on system passivity have been developed. Passivity is a sufficient condition for stability and it has been shown that the interconnection of passive systems is also passive. Because of that, methods based on passivity have gained popularity in teleoperation. One of the most remarkable passivity-based approaches for control of teleoperated systems is Time Domain Passivity Approach (TDPA) \cite{hannaford02,ryu10}. TDPA is less conservative and more robust compared to the other approaches based on passivity. Little knowledge of the teleoperation system or of the external factors (human operator and environment) is needed. In TDPA, the energies flowing in and out of the channel are monitored by a passivity observer (PO), and a passivity controller (PC) acts as an adaptive damper in order to dissipate the necessary amount of energy to ensure passivity of the channel. The adaptive characteristic of TDPA makes it less conservative than the other approaches based on passivity, where the energy dissipating elements are designed for the worst case scenario. However, despite these advantages, TDPA also presents some conservatism, which can affect transparency. Two undesirable effects can be felt when teleoperating robots using TDPA. The first is a mismatch in the perceived task impedance ( see \cite{lawrence93}). The second, which is tackled in this paper, is a position drift between master and slave devices caused by the admittance-type PC on the slave side. \par
Up to now three methods were suggested in order to deal with position drift in TDPA-based teleoperation. In \cite{chawda15} a method to enable position synchronization in power-based TDPA \cite{ye13} based on \textit{r-passivity} \cite{chopra08} was presented. However, due to the conservatism of power-based TDPA, the force reflection is poor and so is the transparency. In energy-based TDPA, a new formulation for the passivity controller was introduced \cite{artigas10}, which, besides removing energy to enforce passivity, also adds energy to compensate for the position drift. The main idea was to emulate a \textit{lossless} network by re-adding the velocity removed by the slave PC as soon as this correction can be applied without compromising the passivity of the system. Later, \cite{chawda14} proposed a new form of performing this task and still keeping the classical PC and the Time Delay Power Network (TDPN) representation presented in \cite{artigas11}. Despite being able to compensate for the position drift, these two methods present a drawback. As mentioned in \cite{chawda14}, force spikes are generated when drift compensation happens after significant position drift has been accumulated. This behavior affects the natural feeling desired for teleoperation since force spikes are suddenly felt by the operator during free-space motion or wall contact. Adding to that, these spikes can put the hardware and the human operator's integrity at risk.\par
In this paper we propose a novel drift compensation method for TDPA-based teleoperation, which presents the following advantages:
\begin{itemize}
\item As in \cite{chawda14}, this approach is also based on TDPA and carries along the advantages of such formulation.
\item Our approach provides smoother drift compensation (with lower forces caused by the compensator), enhancing the natural feeling of free-space motion and wall contact.
\item We provide a way of tuning the behavior of the compensator in order to achieve desired performance (smoothness and compensation speed). 
\end{itemize}
\par
The application of this compensator extends the use of TDPA to tasks where position synchronization is an important requirement. Our approach is able to solve the position-drift issue in TDPA while keeping its original robustness. It also leads to an improvement in what concerns safe human-robot interaction since the forces are kept within the normal range of TDPA-based teleoperation in contrast to the force spikes seen in other drift-compensation methods within TDPA. 

\section{TIME DOMAIN PASSIVITY APPROACH}

Time Domain Passivity Approach \cite{hannaford02,ryu10} is a widely used method to ensure stability of bilateral teleoperation systems. It consists of measuring the energy flow in the system and adaptively dissipating energy in order to enforce passivity. A useful tool to facilitate the application of this approach is to described the entire system in network representation and the communication channel as a Time Delay Power Network (TDPN). The TDPN is a two-port network through which the flow and effort variables are exchanged between master and slave. A TDPN can add time delays, jitter and package losses to the transmitted data, which makes it a suitable representation for the communication channel. \par
\begin{figure}[thpb]
\centering
\includegraphics[trim={1cm 1.2cm 6.5cm 6.7cm},clip,width=1\linewidth]{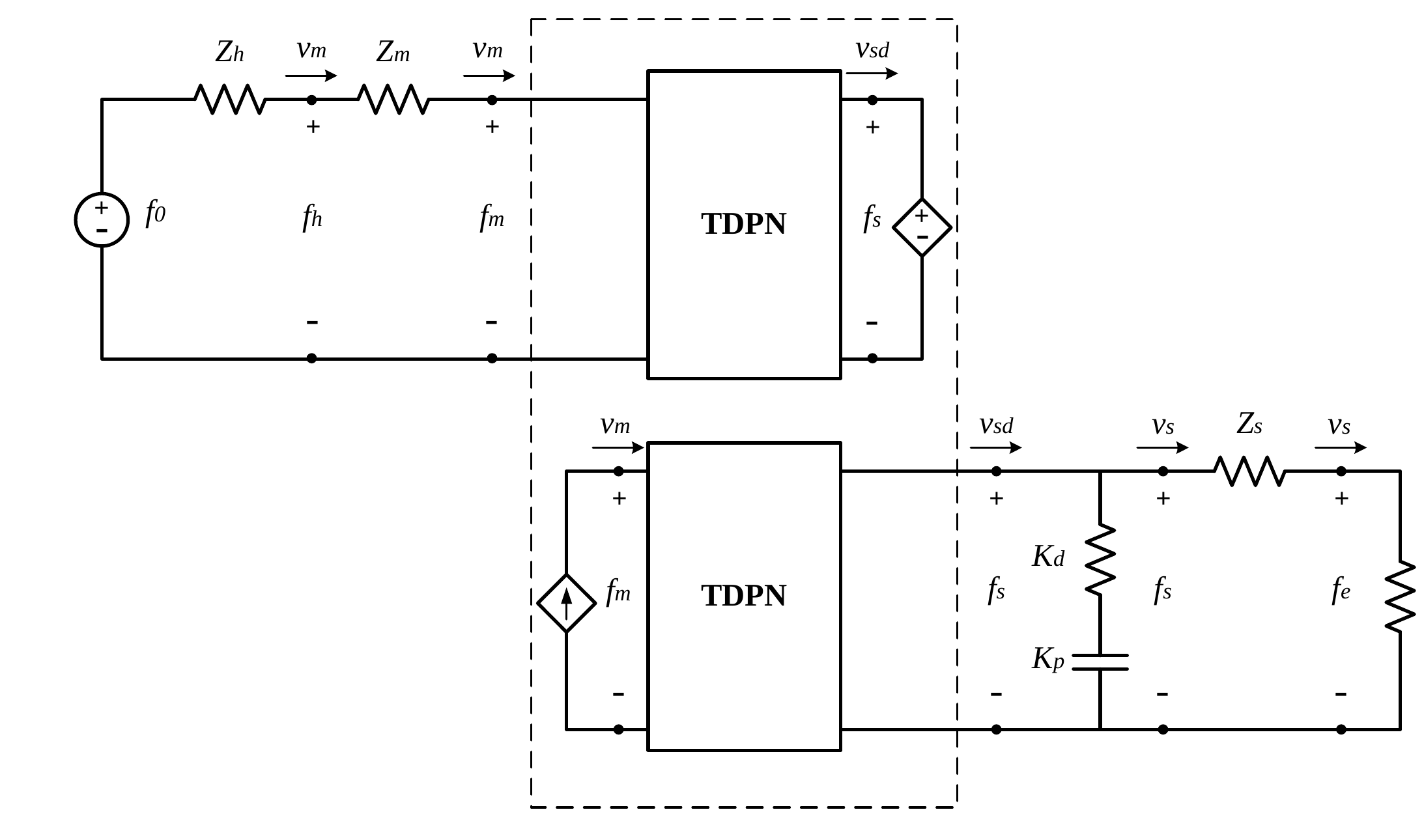}
\caption{Circuit representation of a P-F teleoperation architecture with TDPNs.}
\label{fig:circuit_TDPN}
\end{figure}

The teleoperation scheme analyzed in this paper is a position-force (P-F) architecture. \cite{lawrence93} By representing the communication channel as a TDPN and using mechanical-electrical analogies, this architecture can be described in the framework presented in \cite{artigas11} (see Fig.~\ref{fig:circuit_TDPN}). $v_m$ and $v_s$ are the velocities of the master and slave devices. $f_h$, $f_e$, and $f_s$ are the forces exerted by the human, the environment, and the slave controller, respectively. $v_{sd}$ is the delayed master velocity that serves as reference to the slave, and $f_m$ is the delayed slave computed force applied to the master device. $Z_h$, $Z_m$, $Z_s$, and $Z_e$ are the impedances from the human operator, the master and slave devices, and the environment, respectively. $K_p$ and $K_d$ are the proportional and integral gains of the slave controller, respectively.

As presented in \cite{anderson92}, passivity of a teleoperation two-port network is sufficient to ensure stable bilateral teleoperation. It is also known that the teleoperation system is passive if all its composing subsystems are also passive. This is assumed to be true for the master and slave subsystems, but not for the communication channel, whose passivity has to be enforced. \par
In order to passivate the communication channel, both the master and the slave TDPNs, shown in Fig.~\ref{fig:circuit_TDPN}, have to be passivated. This is achieved by applying TDPA.

\subsection{Energy-based TDPA}
\label{sec:energy_based}

Energy-based TDPA consists of observing the energies on each side of the TDPN with a \textit{passivity observer} (PO), and dissipating part of the energy to make the channel passive with a \textit{passivity controller} (PC).
\par
\begin{figure}[thpb]
\centering
\includegraphics[trim={7.4cm 8cm 9.6cm 7.4cm},clip,width=1\linewidth]{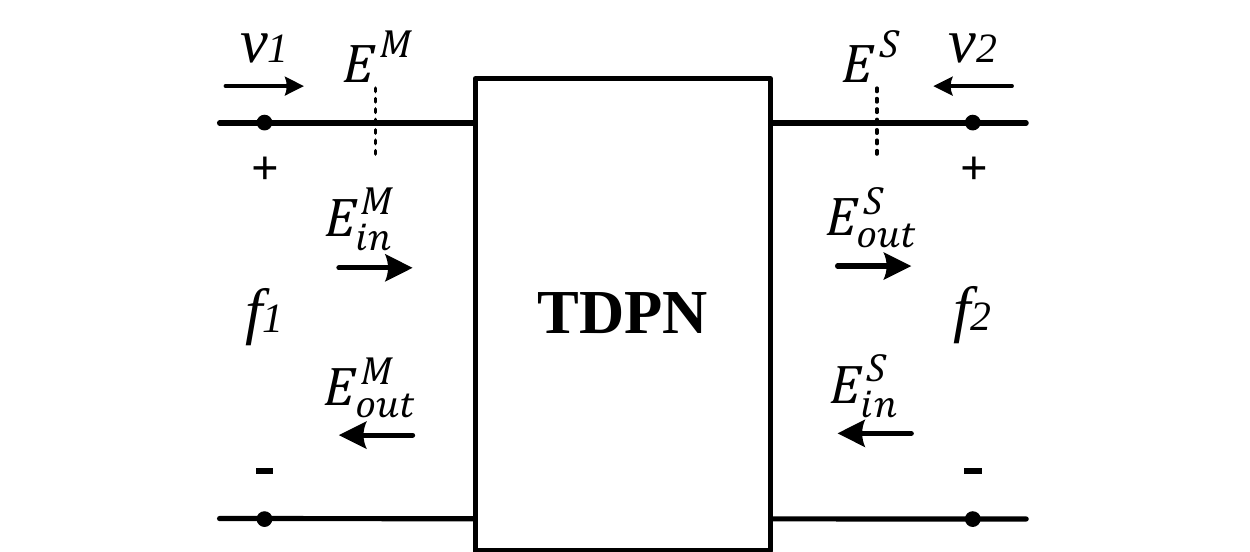}
\caption{Energy flow in the TDPN.}
\label{fig:tdpn_flow}
\end{figure}

The energy flow ($E^N$) in the TDPN (Fig.~\ref{fig:tdpn_flow}) is given by
\begin{equation} \label{eq:energy_flow}
E^N(k)=E^M(k)+E^S(k), \quad  \forall k \geq 0, \\
\end{equation}
where $E^M$ and $E^S$ are the energy contributions from the left and right ports, respectively, which are computed as
\begin{align} \label{eq:energy_integrals}
E^M(k)  = \Delta T \sum_{j=0}^k f_1(j)v_1(j), \\
E^S(k) = \Delta T \sum_{j=0}^k -f_2(j)v_2(j),
\end{align}
where $f_1(k)$ and $v_1(k)$, and $f_2(k)$ and $v_2(k)$ are the sampled forces and velocities checked on the left and right-hand sides of the TDPN, respectively.
\par
The energies on the sides of the TDPN are divided according to the direction of their flow (into or out of the TDPN), as shown in~(\ref{eq:energy_in_out}). 
\begin{equation} 
\begin{split} \label{eq:energy_in_out}
E^M(k) = E^M_{in}(k)-E^M_{out}(k), \quad \forall k \geq 0, \\
E^S(k) = E^S_{in}(k)-E^S_{out}(k), \quad  \forall k \geq 0, 
\end{split} 
\end{equation}
where  $E^M_{in}(k)$, $E^M_{out}(k)$, $ E^S_{in}(k)$, and $E^S_{out}(k)$ are monotonically increasing and non-negative functions, representing the energies flowing in and out of the master and the slave.
\par

A sufficient condition to fulfill the passivity requirement, which can be assessed in case of communication delays is
\begin{align} 
E^{L2R}_{obs}(k) = E^M_{in}(k-T_f(k))-E^S_{out}(k) \geq 0, \quad \forall k \geq 0, \label{eq:energy_obs} \\
E^{R2L}_{obs}(k) = E^S_{in}(k-T_b(k))-E^M_{out}(k) \geq 0, \quad  \forall k \geq 0, \label{eq:energy_obs1}
\end{align} 
where $E^{L2R}_{obs}(k)$ and $E^{R2L}_{obs}(k)$ are the observed left-to-right and right-to-left energy flows observed on the right and left-hand sides of the TDPN, respectively. $T_f(k)$ and $T_b(k)$ are the forward and backward delays, respectively.
\subsubsection{Passivity Observer}

The real-time computation of the observed passivity conditions given by~(\ref{eq:energy_obs}) and (\ref{eq:energy_obs1}) is performed by the POs, which observe the energies $W_M$ and $W_S$ on the master and slave sides, respectively. In order to take the previous dissipation performed by the PC into account, the computations performed by the POs are given by
\begin{align} \label{eq:po}
W_M(k) = E^S_{in}(k-T_b(k))-E^M_{out}(k)+E^M_{PC}(k-1), \\
W_S(k) = E^M_{in}(k-T_f(k))-E^S_{out}(k)+E^S_{PC}(k-1). 
\end{align} 
where $E^M_{PC}(k-1)$ and $E^S_{PC}(k-1)$ are the energies dissipated by the master and slave PCs up to the previous time step. 

\subsubsection{Passivity Controller}

\begin{figure}[thpb]
\centering
\begin{subfigure}[thpb]{1\linewidth}
\includegraphics[trim={4.2cm 6.3cm 8cm 6.7cm},clip,width=1\linewidth]{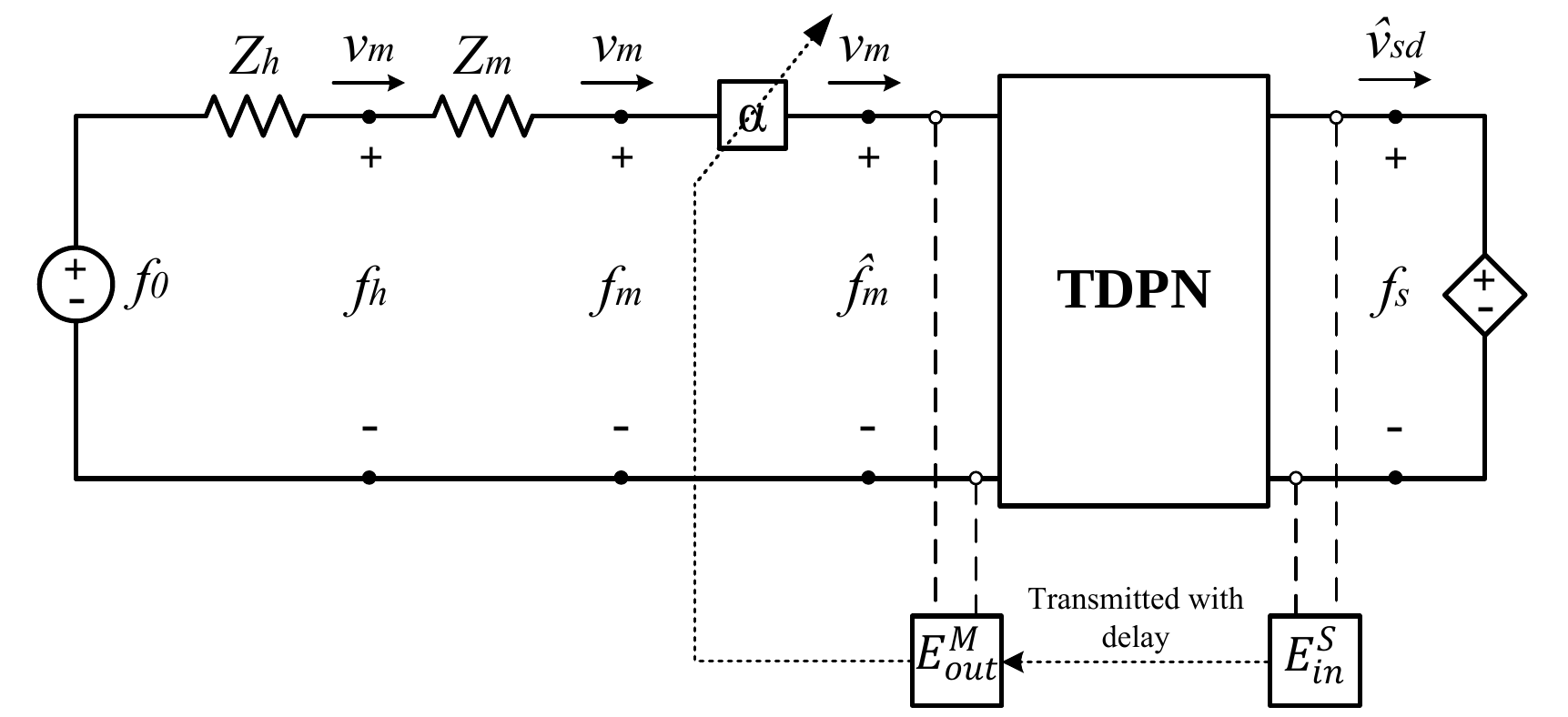}
\caption{PO-PC in impedance configuration.}
\label{fig:impedance}
\end{subfigure}
\begin{subfigure}[thpb]{1\linewidth}
\centering
\includegraphics[trim={6cm 6.7cm 6.5cm 6cm},clip,width=1\linewidth]{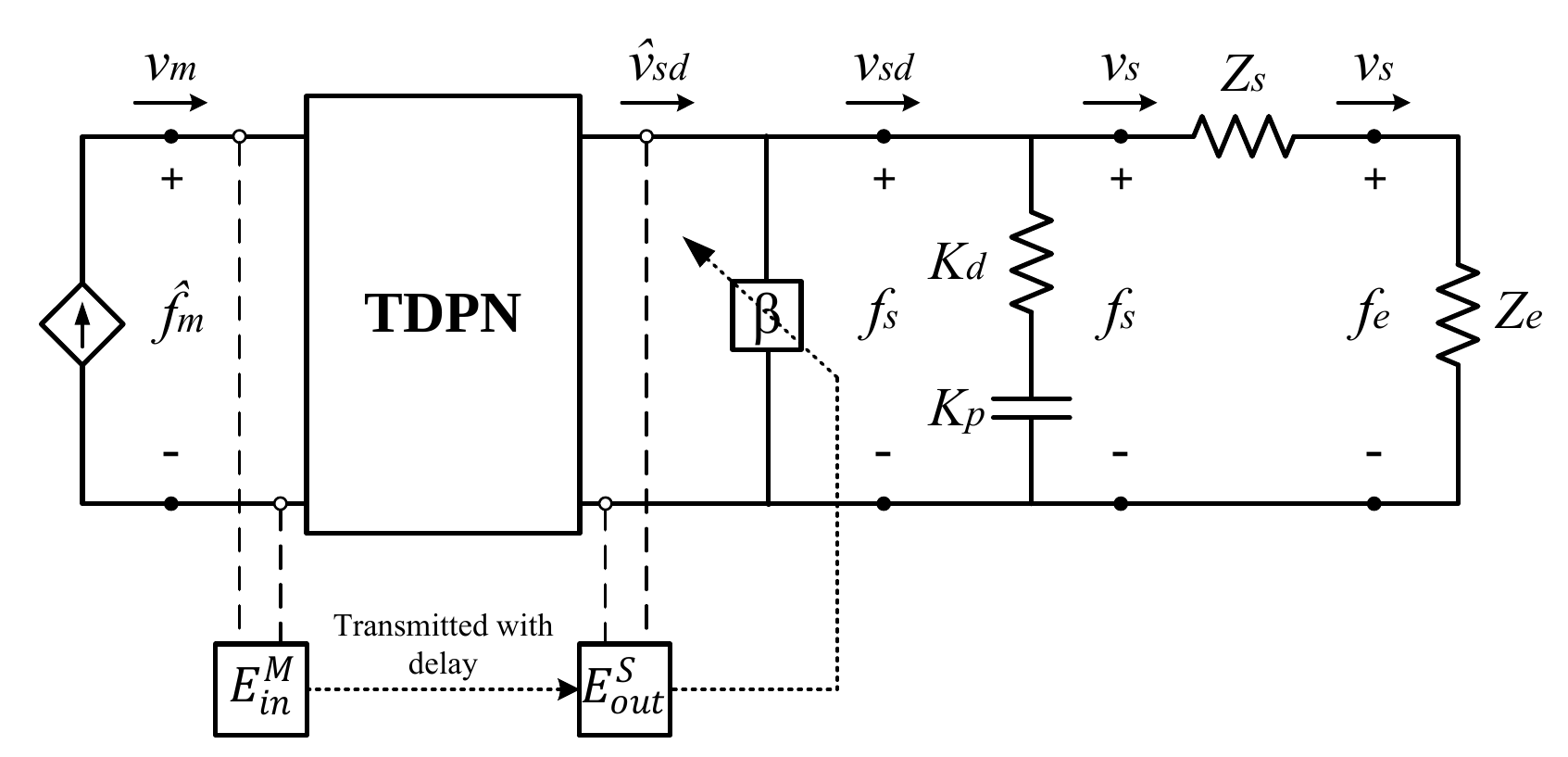}
\caption{PO-PC in admittance configuration.}
\label{fig:admittance}
\end{subfigure}
\caption{P-F architecture with TDPNs passivated with PO-PC.}
\label{fig:po_pc}
\end{figure}

The passivity controllers are adaptive energy-dissipating elements, which are placed at the ports of the TDPN in order to dissipate any excess of energy observed by the POs. Fig.~\ref{fig:po_pc} shows the PC being applied in both impedance (master side) and admittance (slave side) configurations for a P-F architecture with a PD controller on the slave side.

In the admittance type PC (Fig.~\ref{fig:admittance}) on the slave side, the \textit{untouched} velocity signal coming from the master, $\hat{v}_{sd}(k)=v_m(k-T_f(k))$, is modified as follows:
\begin{equation} \label{eq:pc_adm}
v_{sd}(k)=\hat{v}_{sd}(k)+\beta(k)f_s(k)
\end{equation}
$\beta$ is an adaptive energy dissipating element, which can be obtained as
\begin{equation} \label{eq:beta}
\beta(k)=
\begin{cases}
    0                                                        & \text{if } W_S(k)>0\\
    -\cfrac{ W_S(k)}{\Delta Tf^2_s(k)} & \text{else, if } |f_s(k)|>0
\end{cases}
\end{equation}
and the energy dissipated by the PC on the slave side is
\begin{equation}
E^S_{PC}(k)=\Delta T \sum_{j=0}^k\beta(j)f^2_s(j).
\end{equation}

On the master side the PC is applied in impedance configuration and modifies the force coming from the slave in order to keep the system passive.

\subsubsection{Passivity of Ideal Flow and Effort Sources}

As presented in \cite{artigas11}, ideal flow and effort sources can supply and absorb an infinite amount of energy. Therefore, the active energy generated by the TDPN that flows towards the ideal source does not affect the system passivity. For that reason a PO-PC pair is only necessary at the opposite side to the flow and effort sources, as shown in Fig.~\ref{fig:po_pc}.

\subsubsection{Variable Delays}
Due to their non-negative and monotonically increasing characteristic, the values of the input energies $E^M_{in}(k-T_f(k))$ and $E^S_{in}(k-T_b(k))$ are always less than or equal to their non-delayed values ($E^M_{in}(k)$ and $E^S_{in}(k)$), which ensures that the passivity of the system is conserved even with large and variable delays. This is an advantage of energy-based TDPA in contrast with other methods that require the derivative of the time delays, $\dot{T}_f(k)$ and $\dot{T}_b(k)$, to be no greater than one (see \cite{chawda15}). Please, refer to \cite{ryu10} for a more detailed explanation on the validity of energy-based TDPA in variable-delay and package-loss scenarios.


\subsection{Power-based Passivity Control}
\label{sec:power_based}
In power-based approaches, such as \cite{chawda15} and \cite{ye13}, passivity is achieved by ensuring that the power flow of the channel is positive at all time steps. Given that the energy flow ($E^N$) in the TDPN is the integral of the power flow ($P^N$), 
\begin{equation}
E^N(k)  = \Delta T \sum_{j=0}^k P^N(j),
\label{eq:int_power}
\end{equation}
a sufficient condition to ensure passivity of the channel ($E^N(k)\geq 0, \ \forall k$) is to ensure that $P^N(k)\geq 0$ at all time-steps. Based on that, the PO-PC pairs in power-based passivity approaches are set to monitor the power flow in the channel and dissipate negative power to ensure passivity. \par
As mentioned in \cite{chawda15}, the advantages of power-based methods are smoother force reflection and simpler computation in comparison with energy-based approaches. Nevertheless, these methods provide poorer transparency and lower force reflection due to their conservatism in dissipating any negative power, instead of regarding the energy flow itself. 

\section{PROPOSED APPROACH: SMOOTH POSITION-DRIFT COMPENSATOR}

This section provides an explanation about the cause of position drift in TDPA-based teleoperation as well as the origin of force spikes in the previously developed approaches for drift compensation. Finally, the proposed approach will be described, and it will be shown that position drift can be compensated without having high force spikes. 
 
\subsection{Cause of Position Drift}

From Fig.~\ref{fig:admittance} it can be noted that, when the admittance type PC is activated on the slave side, the velocity signal coming from the master is reduced. At these moments the reference to the slave controller ($v_{sd}$) differs from the \textit{untouched} velocity signal coming from the master ($\hat{v}_{sd}$). As soon as the PC becomes inactive, $v_{sd}$ assumes the same value as $\hat{v}_{sd}$ and the velocities can be synchronized. The problem, however, arises when position synchronization is desired, which is the case for most teleoperation applications. Position signals are usually not transmitted through the channel due to limited bandwidth and to the fact that the power conjugate variable for force is velocity. In order to obtain information about the position of the master, the velocity signal has to be integrated on the slave side. However, in order to keep the system passive the velocity signal used on the slave side is $v_{sd}$ and the position command from the master is obtained as
\begin{equation}
x_{sd}=\Delta T\sum_{j=0}^k v_{sd}(j).
\end{equation}
The correspondence between $x_{sd}$ and the actual delayed master device's position, $\hat{x}_{sd}$, is compromised as soon as the PC becomes active for the first time, and the error between them is accumulated whenever the PC modifies the velocity signal. This error remains even when the PC is not active, due to the accumulating characteristic of the integral. The drift between delayed master position and slave reference position is given as
\begin{equation} \label{eq:x_err}
x_{err}(k) = \Delta T\sum_{j=0}^k \hat{v}_{sd}(j) - \Delta T\sum_{j=0}^k v_{sd}(j). 
\end{equation}
By substituting (\ref{eq:pc_adm}) into (\ref{eq:x_err}) we get
\begin{equation} \label{eq:x_err_pc}
 x_{err}(k)= \Delta T\sum_{j=_0}^k  \beta(k)f_s(k).
\end{equation}

From (\ref{eq:x_err_pc}) it can be noted that $x_{err}$ is the integral of the velocity removed by the PC at all time steps. 
\subsection{Origin of Force Spikes}

Up to now, two compensators have been developed in order to eliminate the position drift in energy-based TDPA without decreasing the transparency of the task. In \cite{artigas10} a modification to the classical PO-PC formulation was suggested in order to add extra energy whenever there is a positive gap between the delayed energy flowing into the channel on the master side and the energy flowing out of the channel on the slave side. An algorithm was proposed to keep track of the position error shown in (\ref{eq:x_err}) and to divide it by the sampling period and add it to the velocity coming from the master whenever a passivity gap showed up. By doing that, the error between $x_{sd}$ and $\hat{x}_{sd}$ is liquidated if sufficient gaps appear. \par
\begin{figure}[thpb]
\centering
\includegraphics[trim={6cm 6.7cm 5cm 6cm},clip,width=1\linewidth]{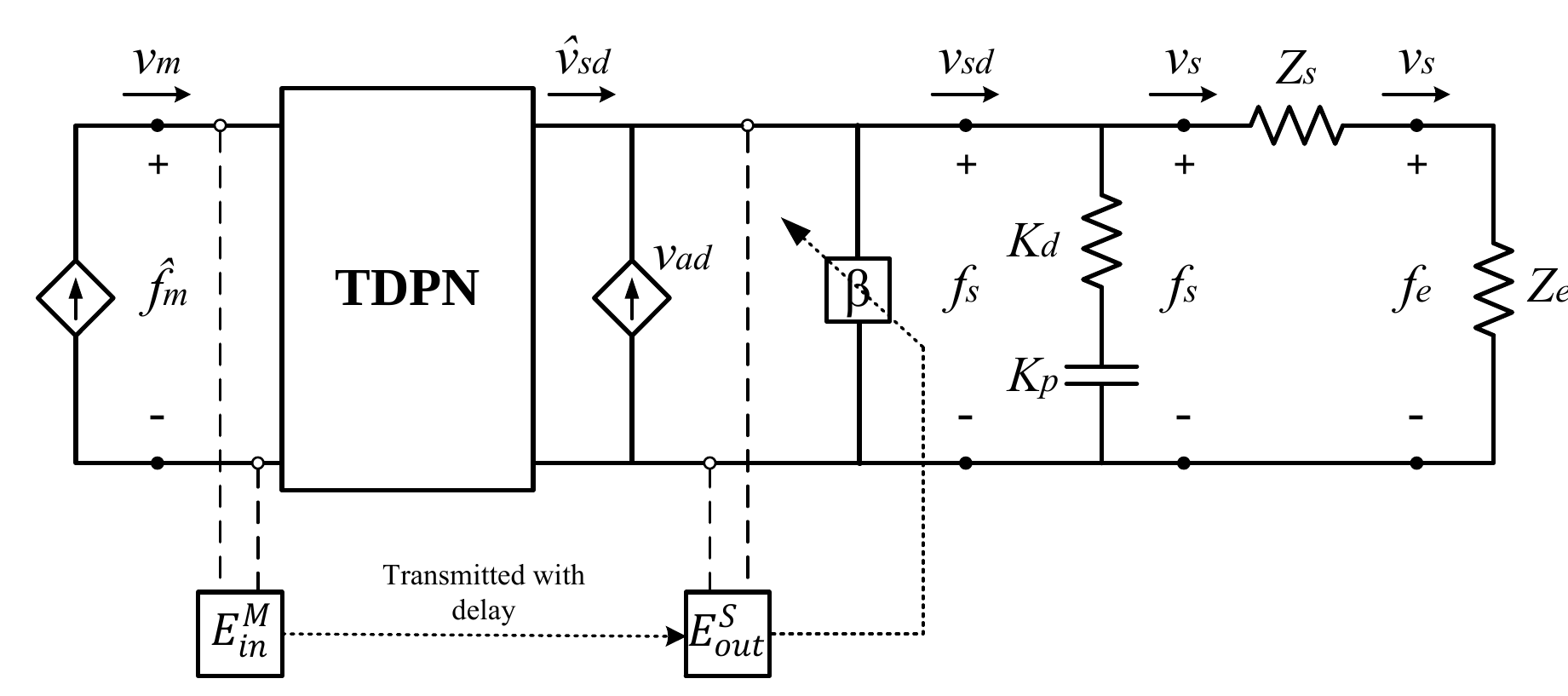}
\caption{Drift compensator on slave side. The compensator is represented by the current source $v_{ad}$. Note that the master side remains the same.}
\label{fig:drift_comp}
\end{figure}

Based on that idea, \cite{chawda14} proposed an approach to keep the classical PO-PC formulation and have the drift compensation part as an ideal current source between the TDPN and the PC, as shown in Fig.~\ref{fig:drift_comp}. In this approach, the current source represented by $v_{ad}$ is responsible for trying to add the correction to eliminate $x_{err}$ and the classic PC checks the signal composed of $\hat{v}_{sd}+v_{ad}$ in order to dissipate the active energy coming from both the TDPN and the current source. \par
Despite the different construction, both methods present the same results and are usually able to eliminate position drift caused by the PC. However, as mentioned in \cite{chawda14}, this way of compensating drift generates undesirable force spikes, especially after a wall contact. These abrupt forces not only affect the natural feeling of teleoperation, but can also damage the hardware or injure the human operator. \par
The root cause of these forces is that the pair, drift compensator and PC, works as an accumulator. During the period when the PC is active the drift compensator tries to compensate the drift by adding extra velocity. However, until a passivity gap appears, the drift is accumulating and so is the contribution added by the compensator. The correction is usually allowed when changing the direction of motion or releasing the wall. At these moments, the accumulated velocity signal is often added at once and an impulse-like signal is given as reference to the slave, adding a force spike to the controller command. \par
In addition to their abrupt nature, another drawback of those approaches is that the difference between the reference coming from the master and the one given to the slave controller is taken into account and not the actual drift between master and slave, which would be given by $\hat{x}_{sd}-{x}_s$. Correcting the reference signal has been proven an effective way of eliminating drift. However, other methods could be applied in order to avoid force spikes and provide the operator with a smoother feeling of teleoperation.

\subsection{Proposed Drift Compensator}

In order to compensate position drift in a smoother manner, a new approach was developed. Our approach is based on the same architecture as presented in \cite{chawda14} (see Fig.~\ref{fig:drift_comp}). However, a different expression for the velocity source ($v_{ad}$) was developed. The velocity added by the proposed compensator is given as

\begin{equation} \label{eq:comp}
v_{ad}(k)=K\sum_{j=0}^{k-1}\bigl(\hat{v}_{sd}(j)-v_s(j)\bigr).
\end{equation}

From (\ref{eq:comp}) and Fig.~\ref{fig:drift_comp} it can be seen that whenever there is an error between the delayed master position and the slave position, the compensator adds a velocity signal to the \textit{untouched} velocity coming from the master, which is checked by the PO for passivity. As in \cite{chawda14} the PC is responsible for dissipating the extra energy generated by the TDPN or the drift compensator in order to enforce passivity. The gain $K$ from (\ref{eq:comp}) can be appropriately tuned in order to obtain the desired performance. Higher values of $K$ increase the compensation signal and decrease the number of time steps necessary to remove drift. Lower values of $K$ yield lower $v_{ad}$, which compensates the drift in a smoother manner over a larger number of time steps. A gain $K$ of 1 would cause force peaks like the ones generated by the method described in \cite{chawda14}. Future work will include an adaptive way of tuning $K$, but based on the results of \cite{chawda14} it is advisable to keep $0<K<1$. \par 
Due to its integral action, the proposed compensator is able to make the position drift caused by TDPA converge to zero during passivity gaps. During a teleoperation task where TDPA is used together with the proposed compensator, position drift will increase when the PC is active and decrease during passivity gaps. If enough enough passivity gaps appear, the compensator will be able to keep the drift at zero most of the time. This will be further discussed in Section~\ref{sec:exp}.
\par
Another characteristic of this compensator is that it acts together with the slave controller. Whenever there is a passivity gap, the compensator modifies the reference in order to reduce the position error between master and slave. This behavior contributes to emulating a \textit{lossless} communication channel since it makes the values of $E_{out}$ be close to $E_{in}$ at almost all time steps.
\par
The proposed method allows for passive teleoperation while providing position synchronization and keeping the forces in their normal range. This is a desired characteristic since force spikes are not always allowed in teleoperation tasks On the other hand, the use of regular TDPA without drift compensation also prevents the successful completion of the tasks since the position drift can get to a point where the slave barely responds to the master's commands. This novel approach is able to provide position synchronization and forces with regular amplitudes, making it possible to successfully complete teleoperation tasks without putting humans and hardware at risk.  
\par
As in \cite{chawda14}, the passivity of this drift compensator is enforced by the PC, which is implemented after the compensator. If the overall system is passive without time delays, then it is also passive when the TDPN and the PO-PC pair are introduced. Adding to that, it is important to note that the proposed compensator has no negative impact on the transparency of the task. Since the maximum value of the reference $v_{sd}$ given to the plant is defined by the value of $E_{in}$. In case the value of $v_{ad}$ is nonzero, but no passivity gap is present, the PC will take stronger action to dissipate the energy that the compensator intended to add. However, the value of $v_{sd}$ will be the same as if no compensator were present. The presence of the compensator does not contribute to the reduction of the value of $v_{sd}$, unless necessary to compensate drift in the other direction.

\section{EXPERIMENTAL EVALUATION}
\label{sec:exp}

\subsection{1-DoF Setup}

\begin{figure}[thpb]
\centering
\includegraphics[width=0.9\linewidth]{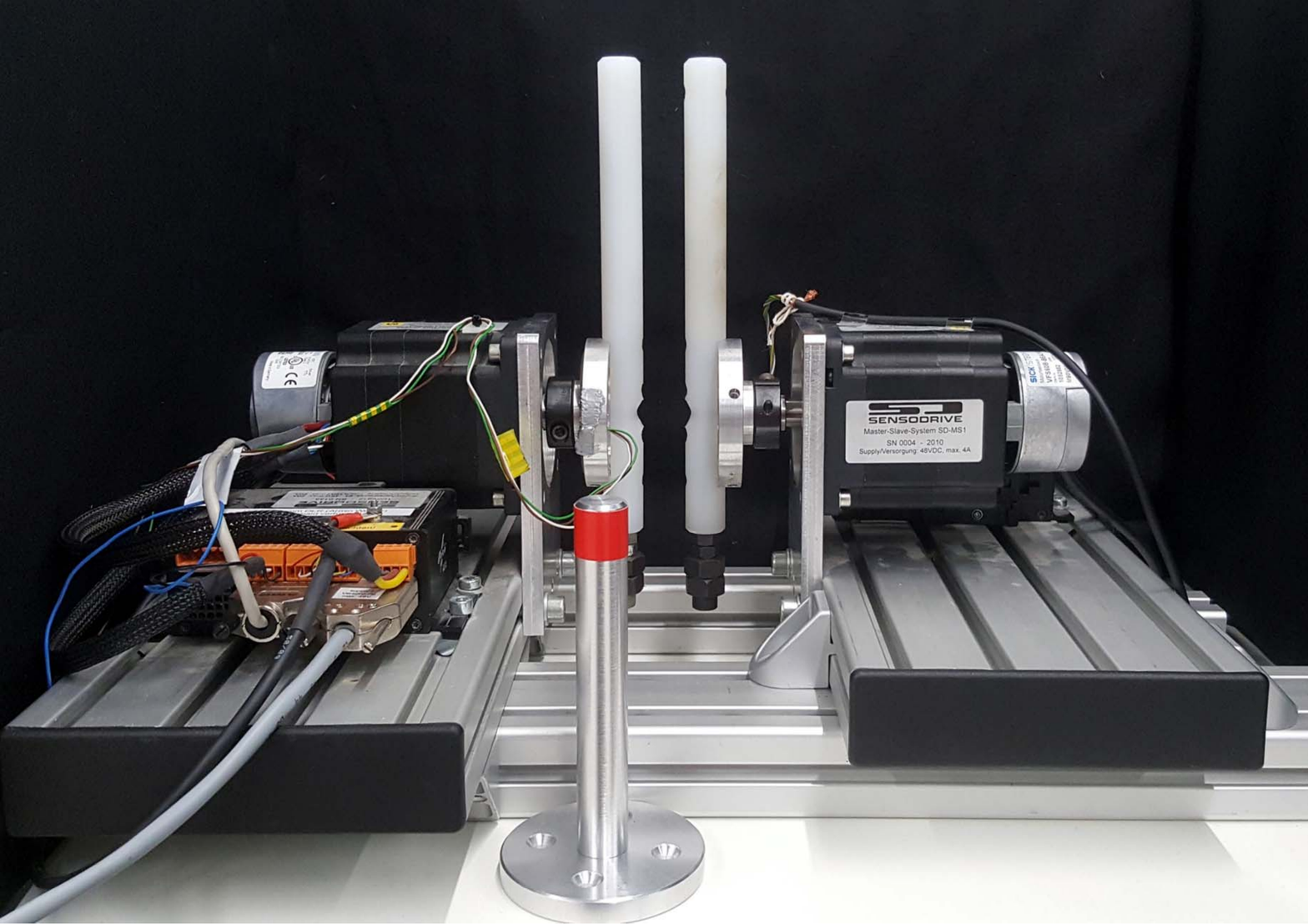}
\caption{1-DoF teleoperation setup used for the experiments.}
\label{fig:setup}
\end{figure}

Experiments were performed using two one-degree-of-freedom (1-DoF) rotational devices (see Fig.~\ref{fig:setup}) composed of two independent motor-gear units, each of them equipped with a torque sensor. The devices were controlled using the same computer, which was running at a sampling rate of 1 kHz. The devices were connected through a position-force architecture and communication delay was simulated in software.

\subsection{Comparison with Previous Approaches}
The experiments were conducted with regular TDPA, the compensator proposed in \cite{chawda14}, and the proposed compensator with a gain $K$ of 0.025. The approach from \cite{artigas10} is not tested here since its results are similar to the ones from \cite{chawda14}, despite having different construction. The method presented in \cite{chawda15} is also not included due to space constraints and to the poor transparency offered by power-based passivity control (see Section~\ref{sec:power_based}). \par
Each approach was tested for constant round-trip delays ($T_{rt}=T_f+T_b$) of 200 ms and 500 ms. The human operator attempted to simulate a sinusoidal reference in position while the slave device performed hard-wall contacts. The figures are divided into four subplots: (a) shows the master and slave positions, (b) shows the control torques acting on the master and the slave, (c) shows the input energy from the slave and the output energy at the master, (d) shows the input energy from the master, the output energy at the slave, and the energy added by the compensators before the PC ($E_{ad}$), except for the regular-TDPA case. \par
\begin{figure}[thpb]
\centering
\begin{subfigure}[thpb]{0.5\linewidth}
\includegraphics[trim={0.2cm 0cm 1.1cm 0.6cm},clip,width=1.5in]{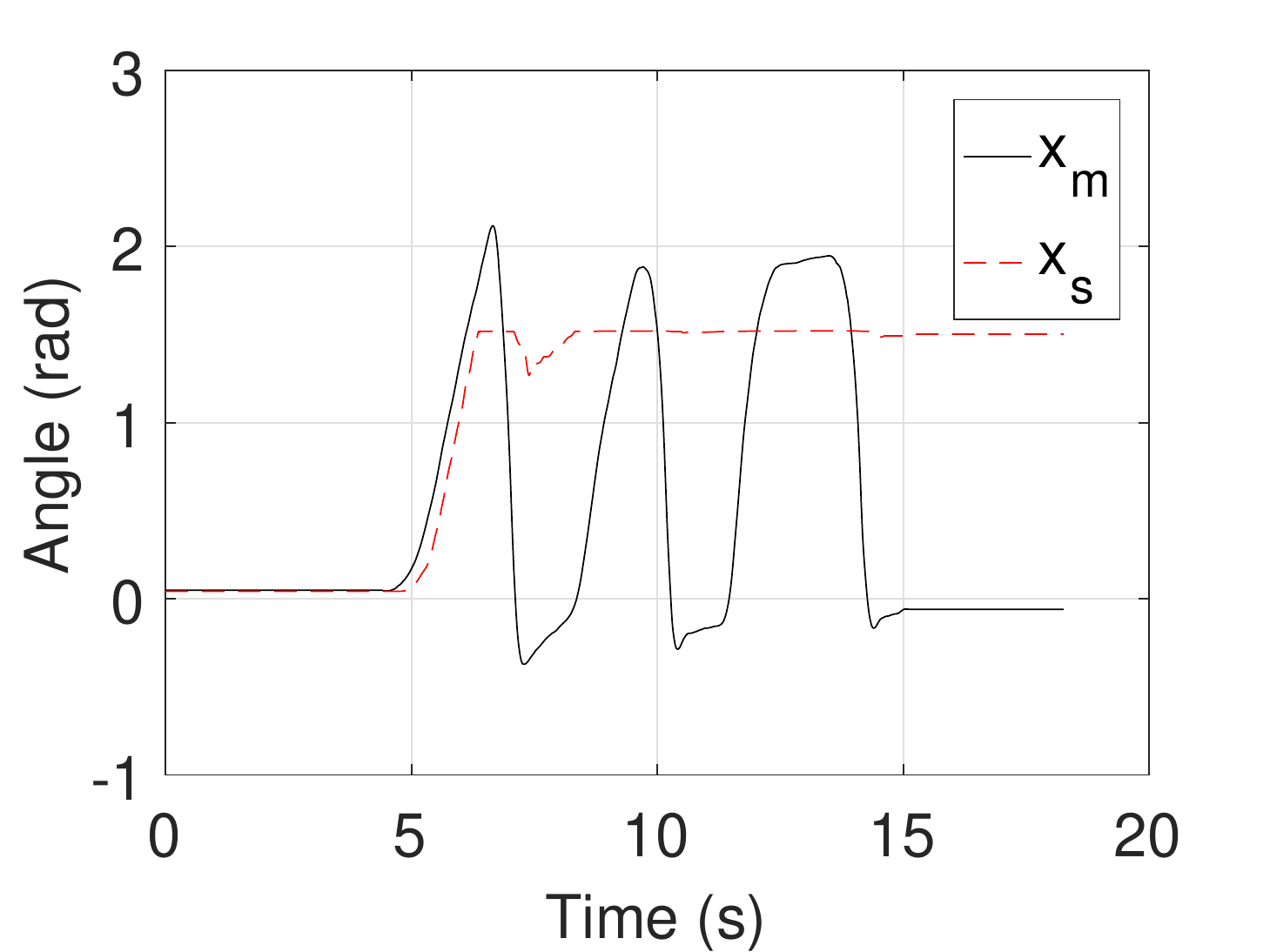}
\caption{}
\label{fig:tdpa200_1}
\end{subfigure}%
~
\begin{subfigure}[thpb]{0.5\linewidth}
\includegraphics[trim={0.2cm 0cm 1.1cm 0.6cm},clip,width=1.5in]{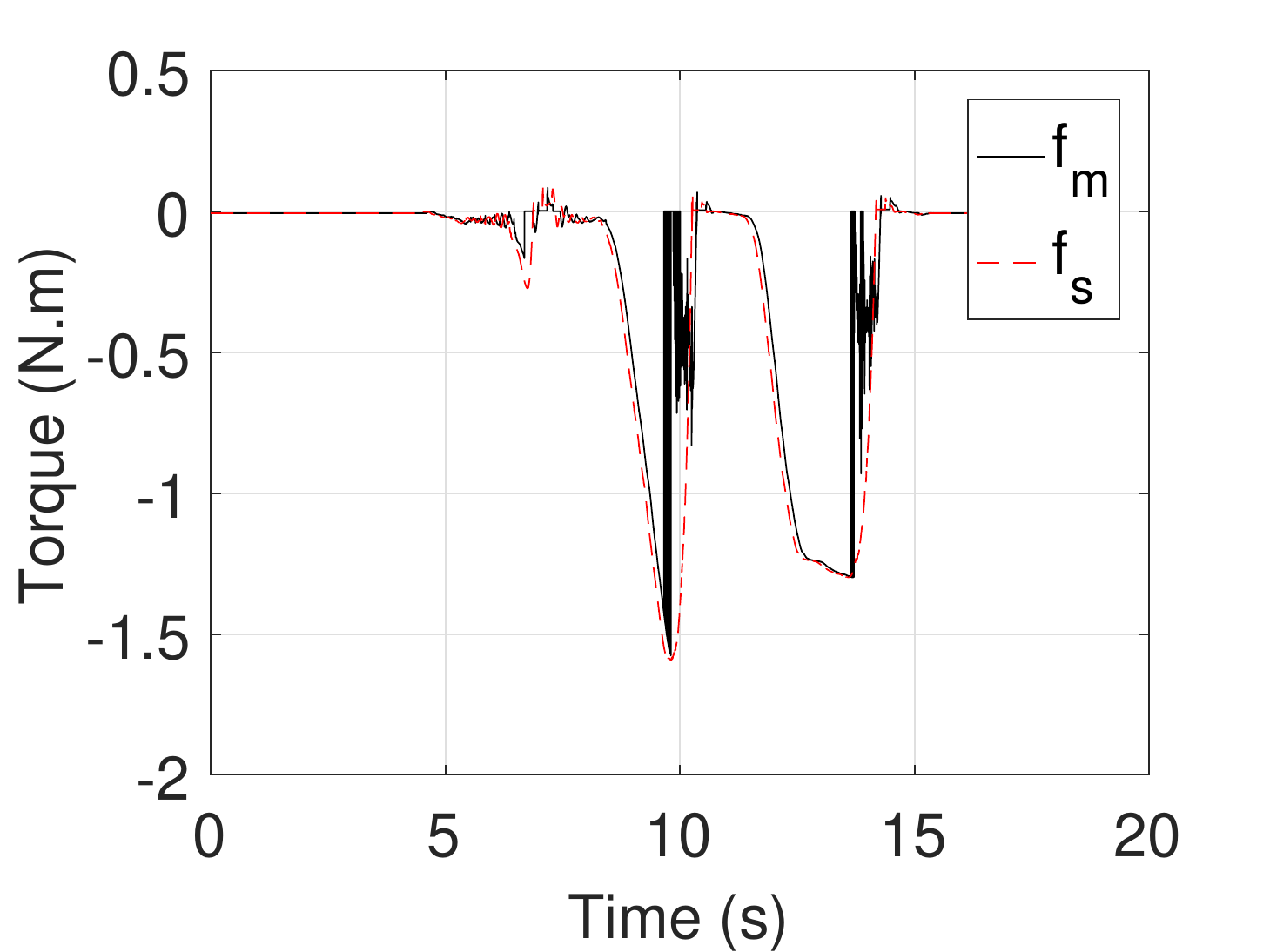}
\caption{}
\label{fig:tdpa200_2}
\end{subfigure}

\begin{subfigure}[thpb]{0.5\linewidth}
\includegraphics[trim={0.2cm 0cm 1.1cm 0.6cm},clip,width=1.5in]{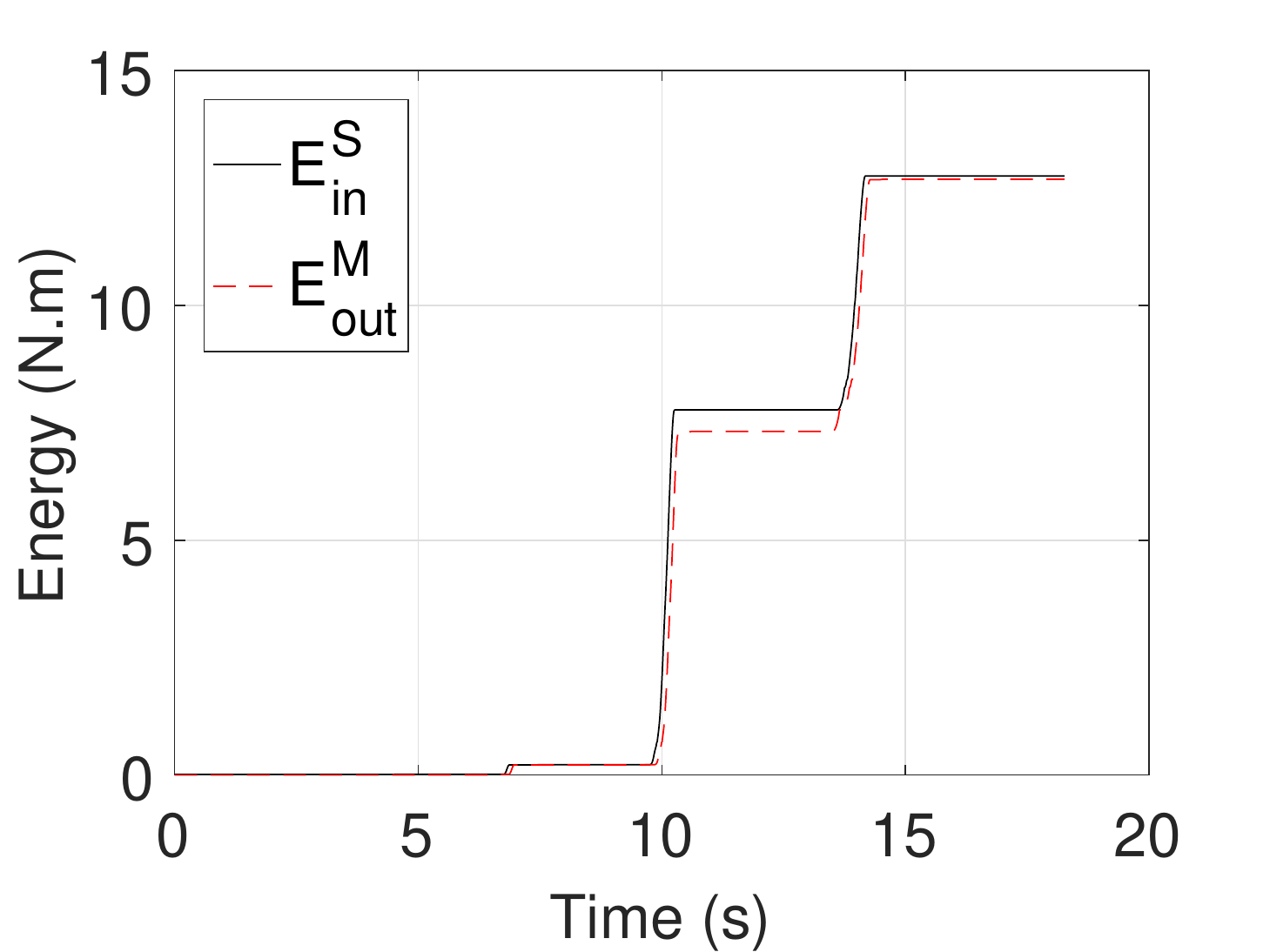}
\caption{}
\label{fig:tdpa200_3}
\end{subfigure}%
~
\begin{subfigure}[thpb]{0.5\linewidth}
\includegraphics[trim={0.2cm 0cm 1.1cm 0.6cm},clip,width=1.5in]{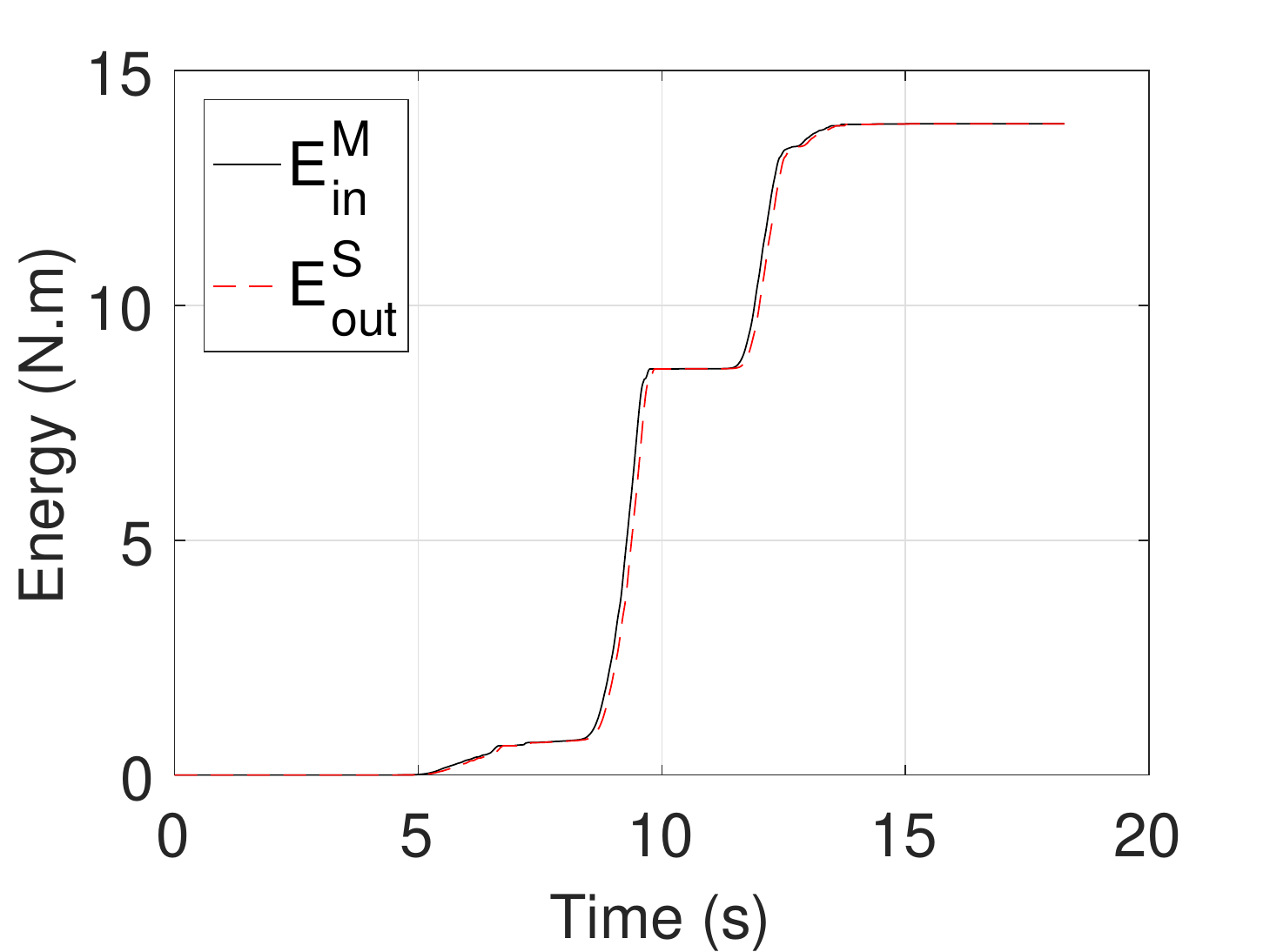}
\caption{}
\label{fig:tdpa200_4}
\end{subfigure}
\caption{$T_{rt}=200$ ms wall contact -- no drift compensator}
\label{fig:tdpa200}
\end{figure}
\begin{figure}[thpb]
\centering
\begin{subfigure}[thpb]{0.5\linewidth}
\includegraphics[trim={0.2cm 0cm 1.1cm 0.6cm},clip,width=1.5in]{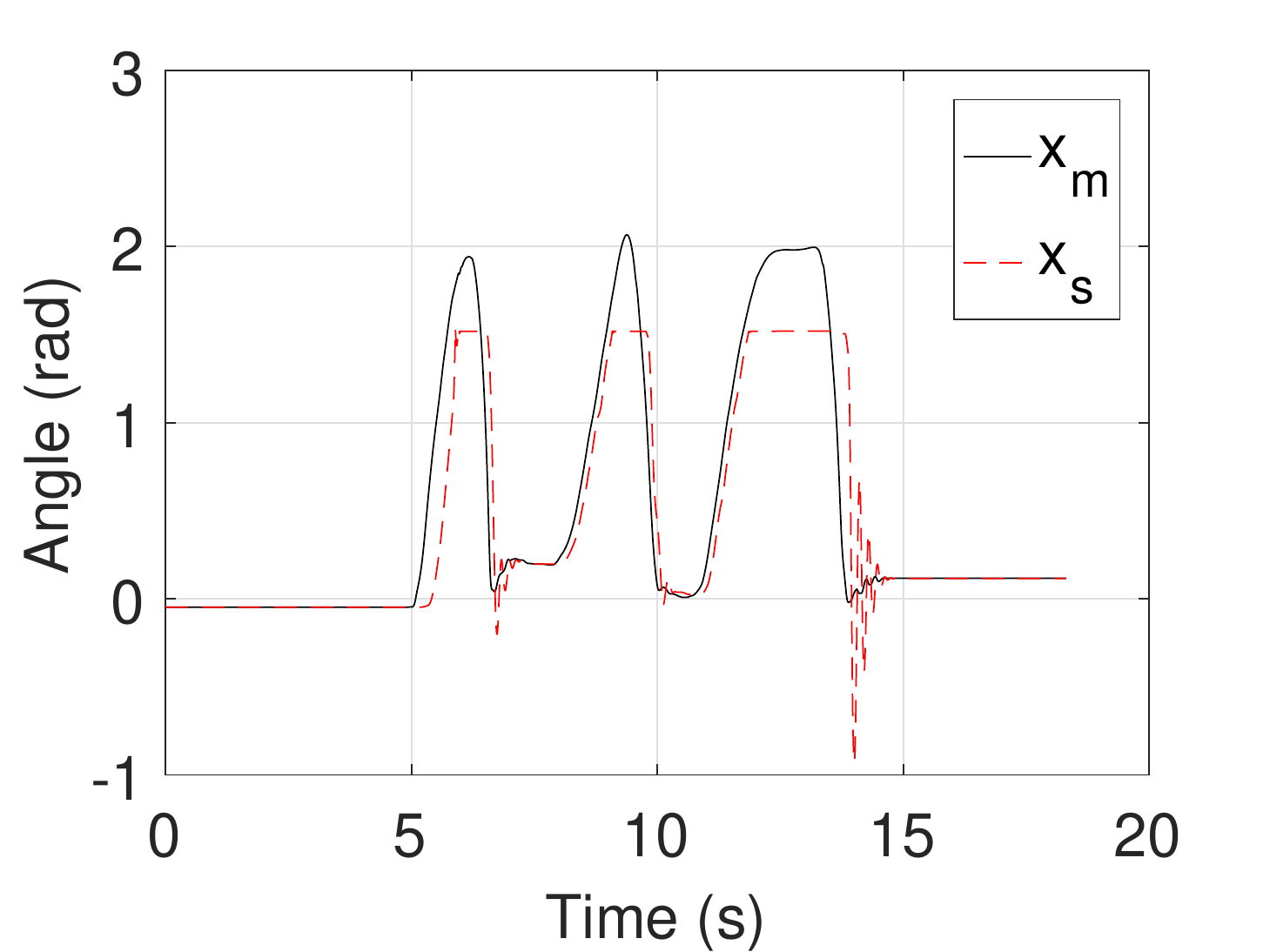}
\caption{}
\label{fig:chawda200_1}
\end{subfigure}%
~
\begin{subfigure}[thpb]{0.5\linewidth}
\includegraphics[trim={0.2cm 0cm 1.1cm 0.6cm},clip,width=1.5in]{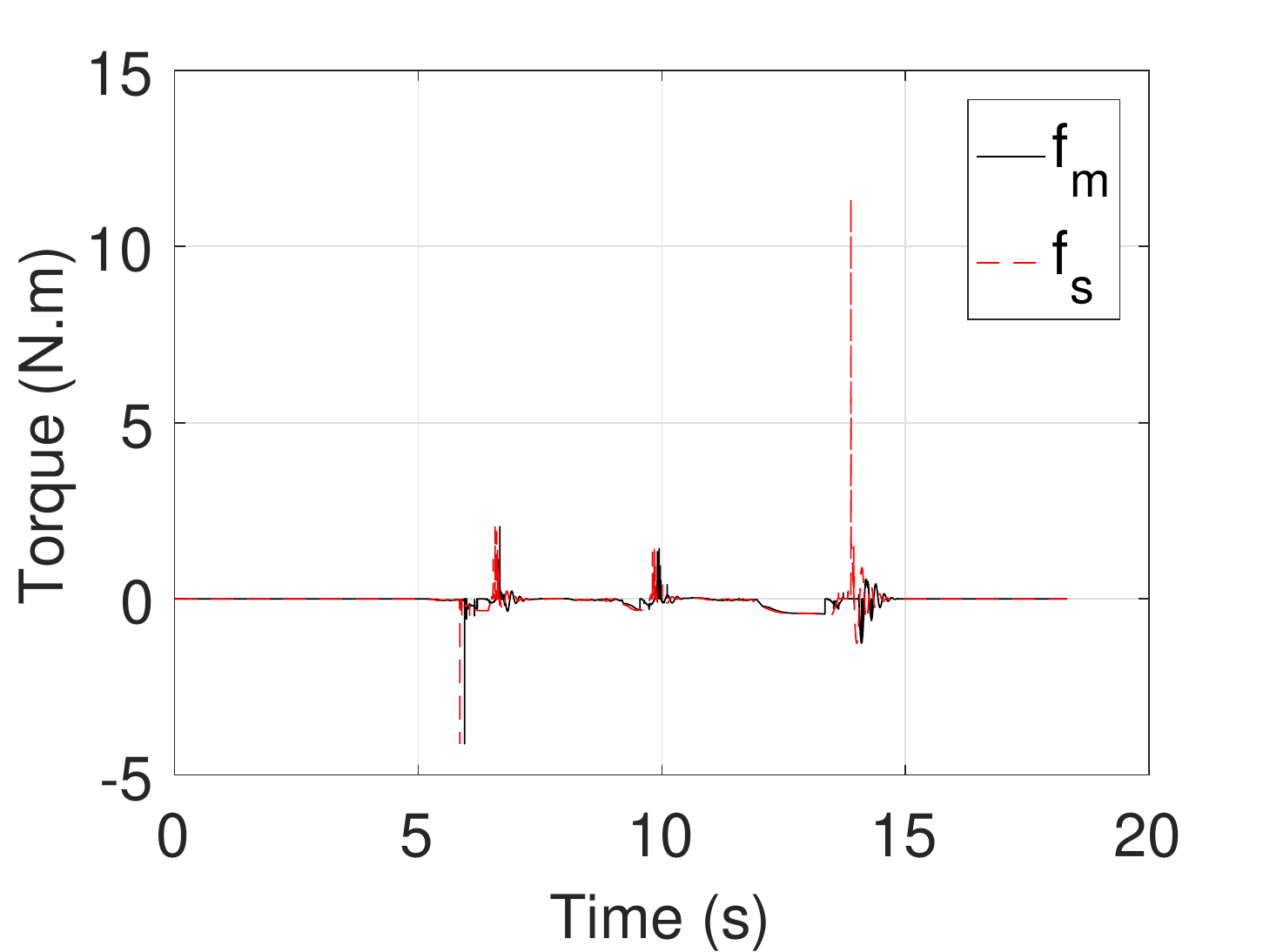}
\caption{}
\label{fig:chawda200_2}
\end{subfigure}

\begin{subfigure}[thpb]{0.5\linewidth}
\includegraphics[trim={0.2cm 0cm 1.1cm 0.6cm},clip,width=1.5in]{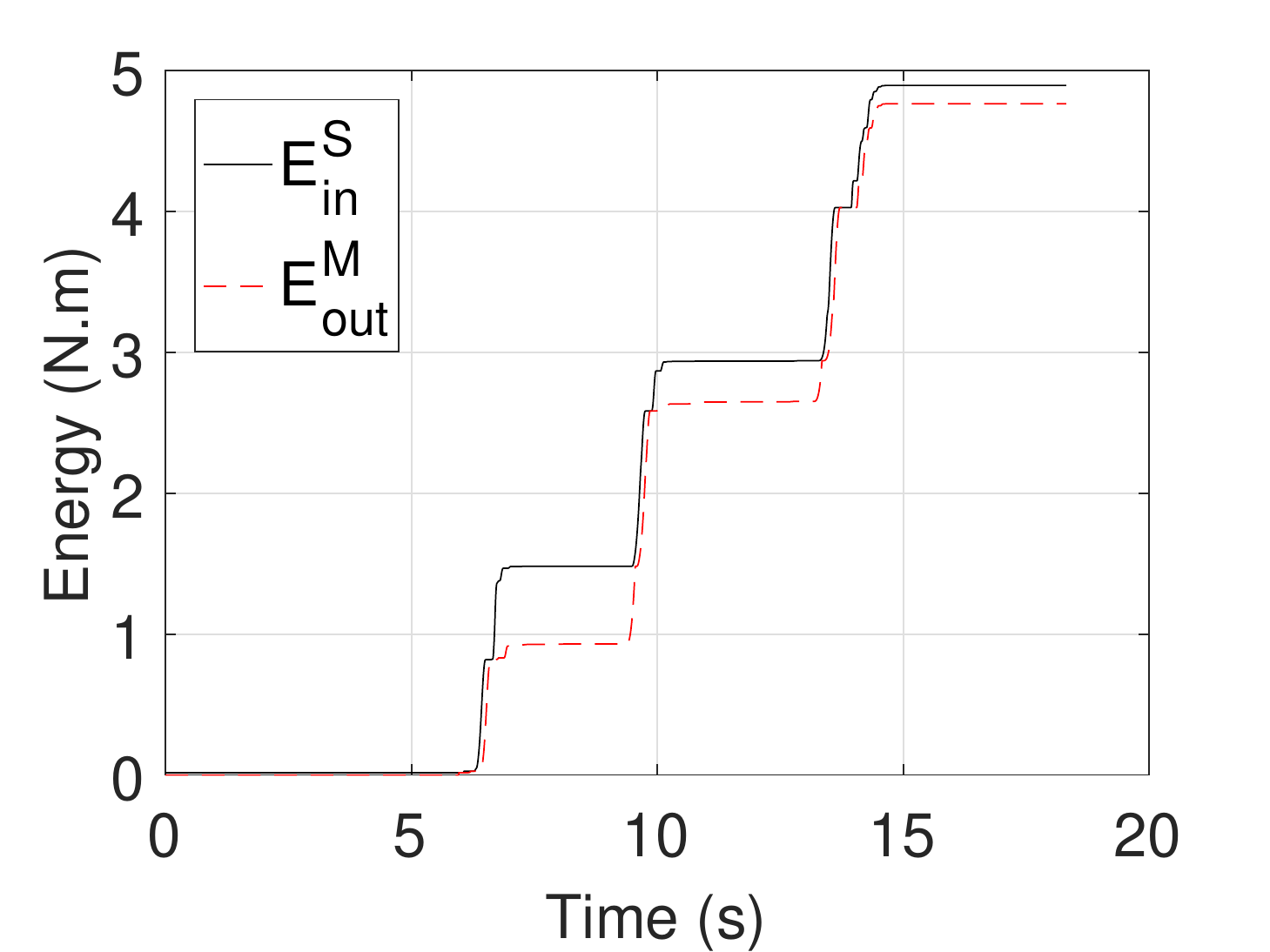}
\caption{}
\label{fig:chawda200_3}
\end{subfigure}%
~
\begin{subfigure}[thpb]{0.5\linewidth}
\includegraphics[trim={0.2cm 0cm 1.1cm 0.6cm},clip,width=1.5in]{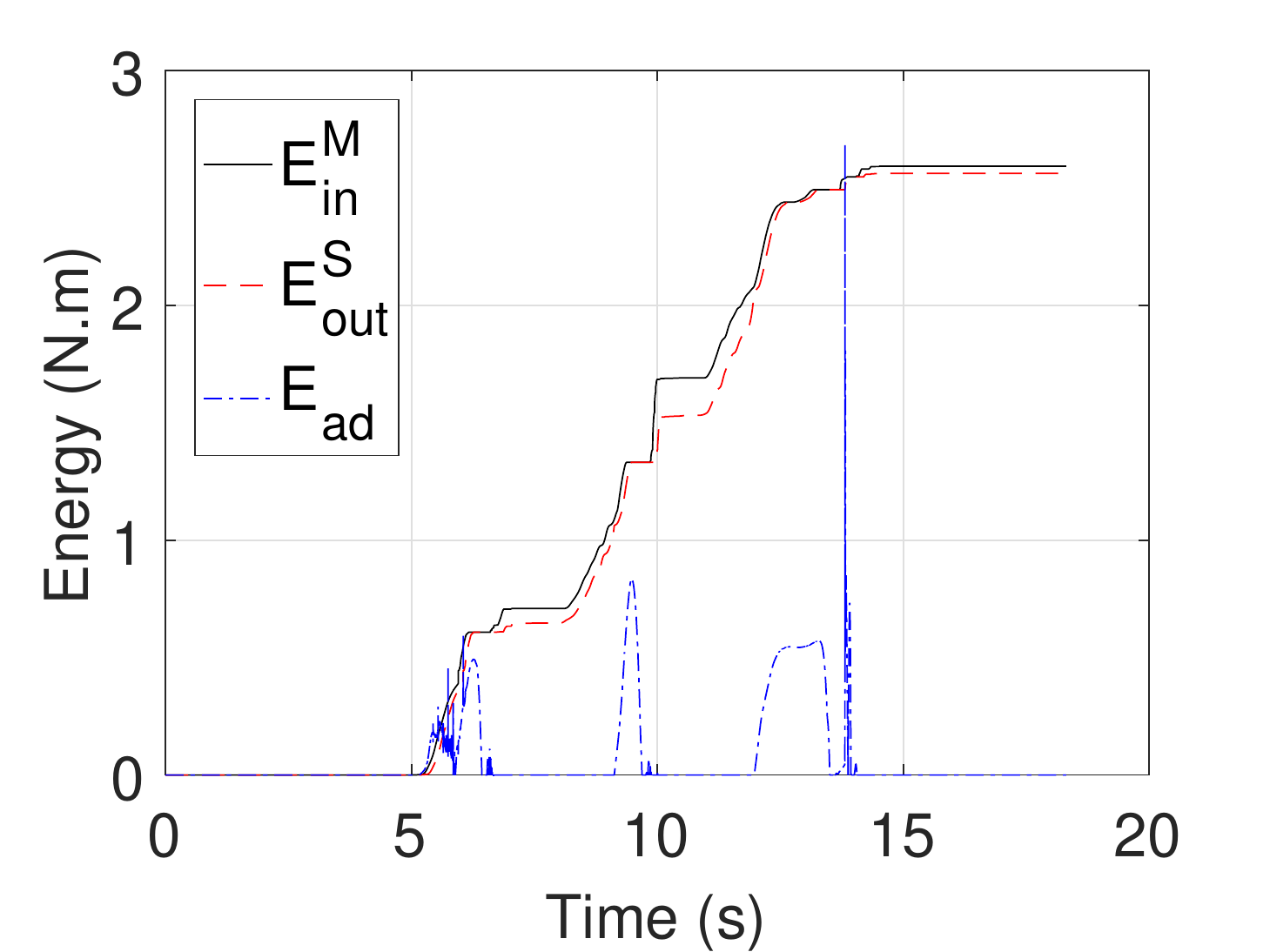}
\caption{}
\label{fig:chawda200_4}
\end{subfigure}
\caption{$T_{rt}=200$ ms wall contact -- compensator from \cite{chawda14}}
\label{fig:chawda200}
\end{figure}
\begin{figure}[thpb]
\centering
\begin{subfigure}[thpb]{0.5\linewidth}
\includegraphics[trim={0.2cm 0cm 1.1cm 0.6cm},clip,width=1.5in]{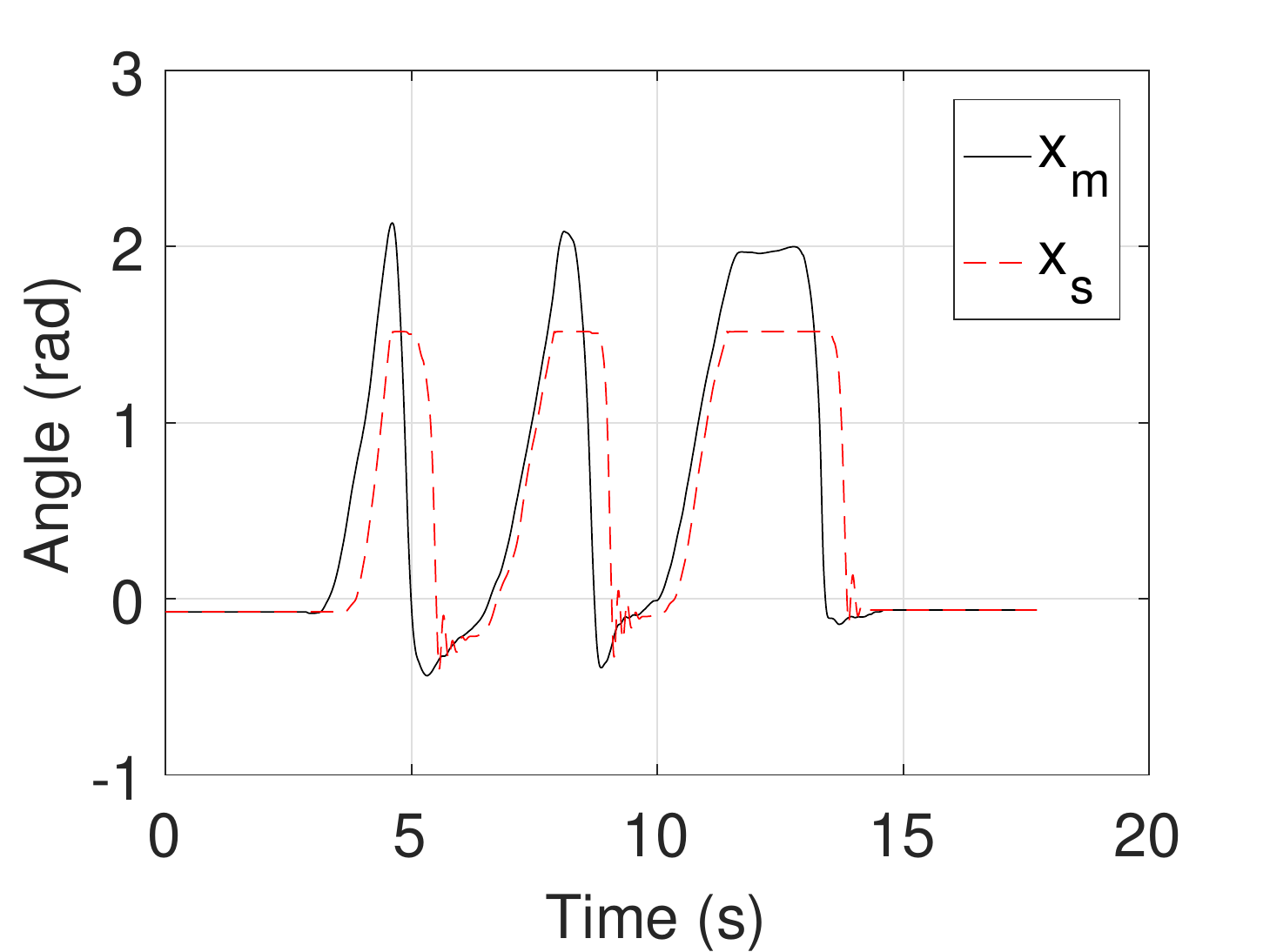}
\caption{}
\label{fig:andre200_1}
\end{subfigure}%
~
\begin{subfigure}[thpb]{0.5\linewidth}
\includegraphics[trim={0.2cm 0cm 1.1cm 0.6cm},clip,width=1.5in]{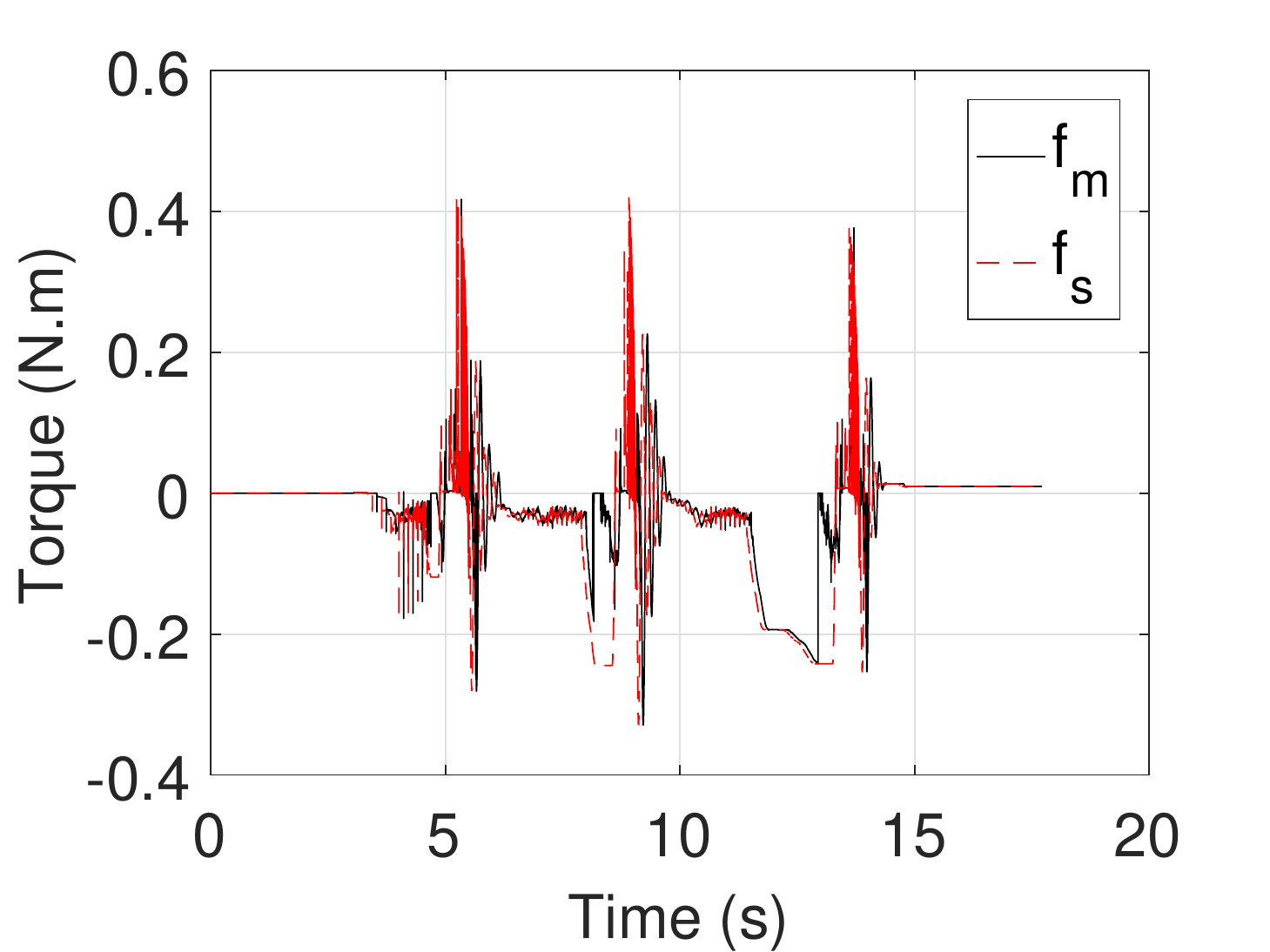}
\caption{}
\label{fig:andre200_2}
\end{subfigure}

\begin{subfigure}[thpb]{0.5\linewidth}
\includegraphics[trim={0.2cm 0cm 1.1cm 0.6cm},clip,width=1.5in]{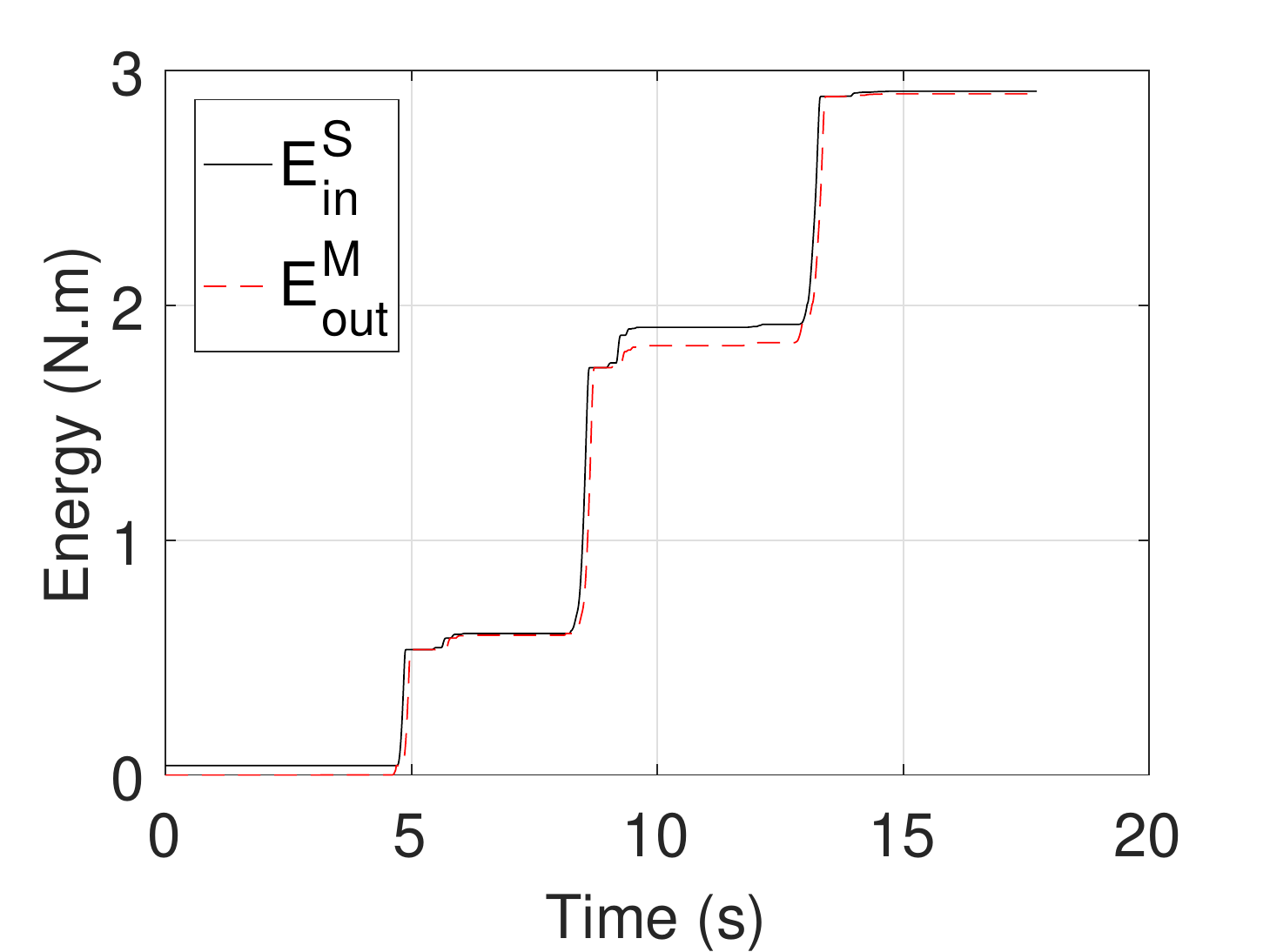}
\caption{}
\label{fig:andre200_3}
\end{subfigure}%
~
\begin{subfigure}[thpb]{0.5\linewidth}
\includegraphics[trim={0.2cm 0cm 1.1cm 0.6cm},clip,width=1.5in]{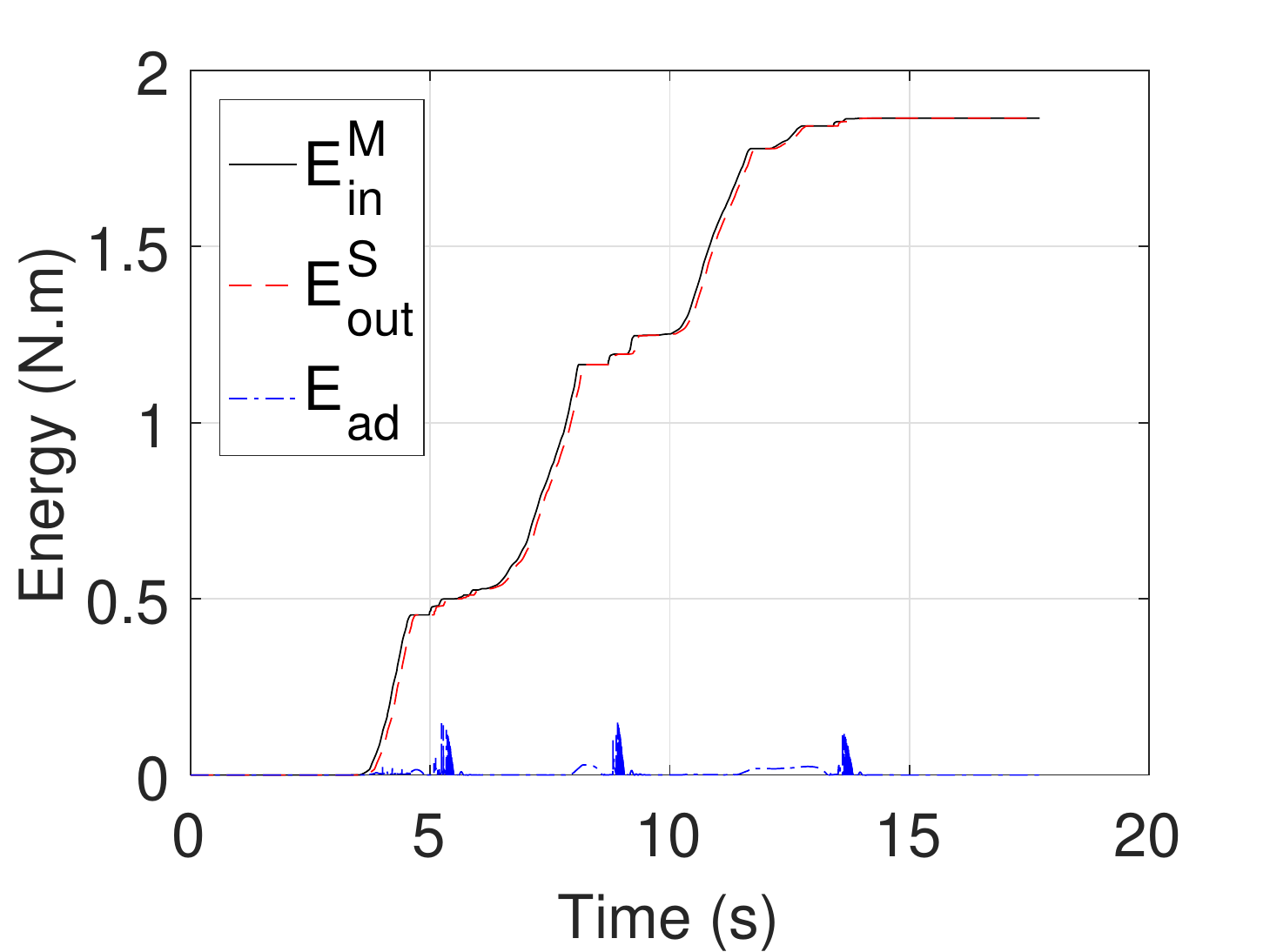}
\caption{}
\label{fig:andre200_4}
\end{subfigure}
\caption{$T_{rt}=200$ ms wall contact -- proposed compensator}
\label{fig:andre200}
\end{figure}

Figs.~\ref{fig:tdpa200}, \ref{fig:chawda200}, and \ref{fig:andre200} show the results of the experiments for a constant delay of 200 ms round-trip. In Fig.~\ref{fig:tdpa200_1} it can be seen that, with regular TDPA, the position offset was increased to a point where the slave barely moved, despite the commands from the master. Due to the drift, the wall contacts happened earlier at each time and the operator had to exert higher forces (Fig.~\ref{fig:tdpa200_2}) in order to follow the sinusoidal reference. With the compensator from \cite{chawda14}, the drift is successfully compensated (Fig.~\ref{fig:chawda200_1}). Nevertheless, oscillations on the slave device are induced by the torque spikes (Fig.~\ref{fig:chawda200_2}) generated when the energy stored by the compensator is released.  It can be noted that the proposed compensator (Fig.~\ref{fig:andre200}) is capable of removing the position drift while causing much smaller oscillations (Fig.~\ref{fig:andre200_1}) and keeping the forces within their normal range (Fig.~\ref{fig:andre200_1}). Moreover, the energy added ($E_{ad}$) by the proposed compensator (Fig.~\ref{fig:andre200_4}) is much lower than the energy added by the compensator from \cite{chawda14} (Fig.~\ref{fig:chawda200_4}).

\begin{figure}[thpb]
\centering
\begin{subfigure}[thpb]{0.5\linewidth}
\includegraphics[trim={0.2cm 0cm 1.1cm 0.6cm},clip,width=1.5in]{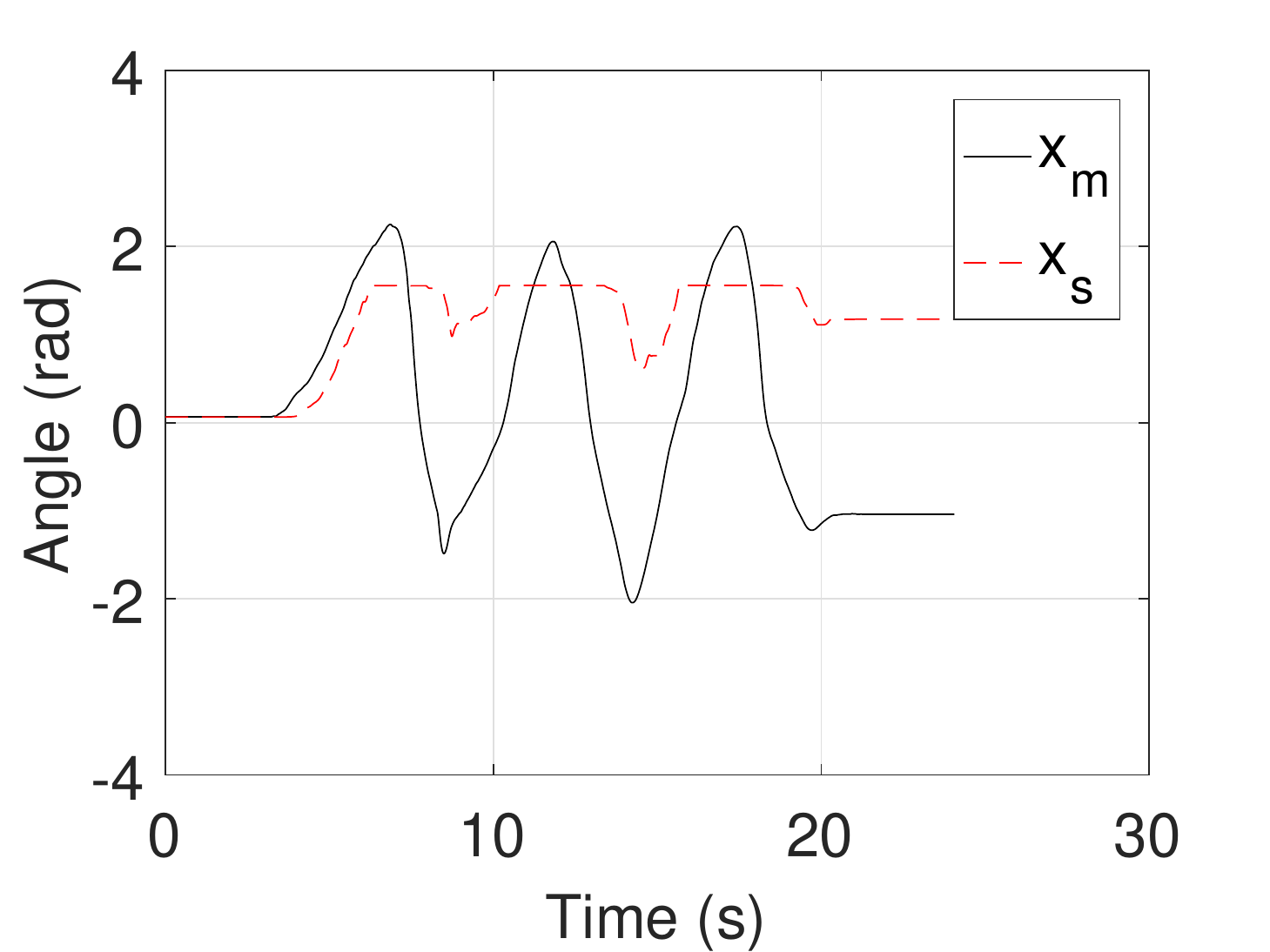}
\caption{}
\label{fig:tdpa500_1}
\end{subfigure}%
~
\begin{subfigure}[thpb]{0.5\linewidth}
\includegraphics[trim={0.2cm 0cm 1.1cm 0.6cm},clip,width=1.5in]{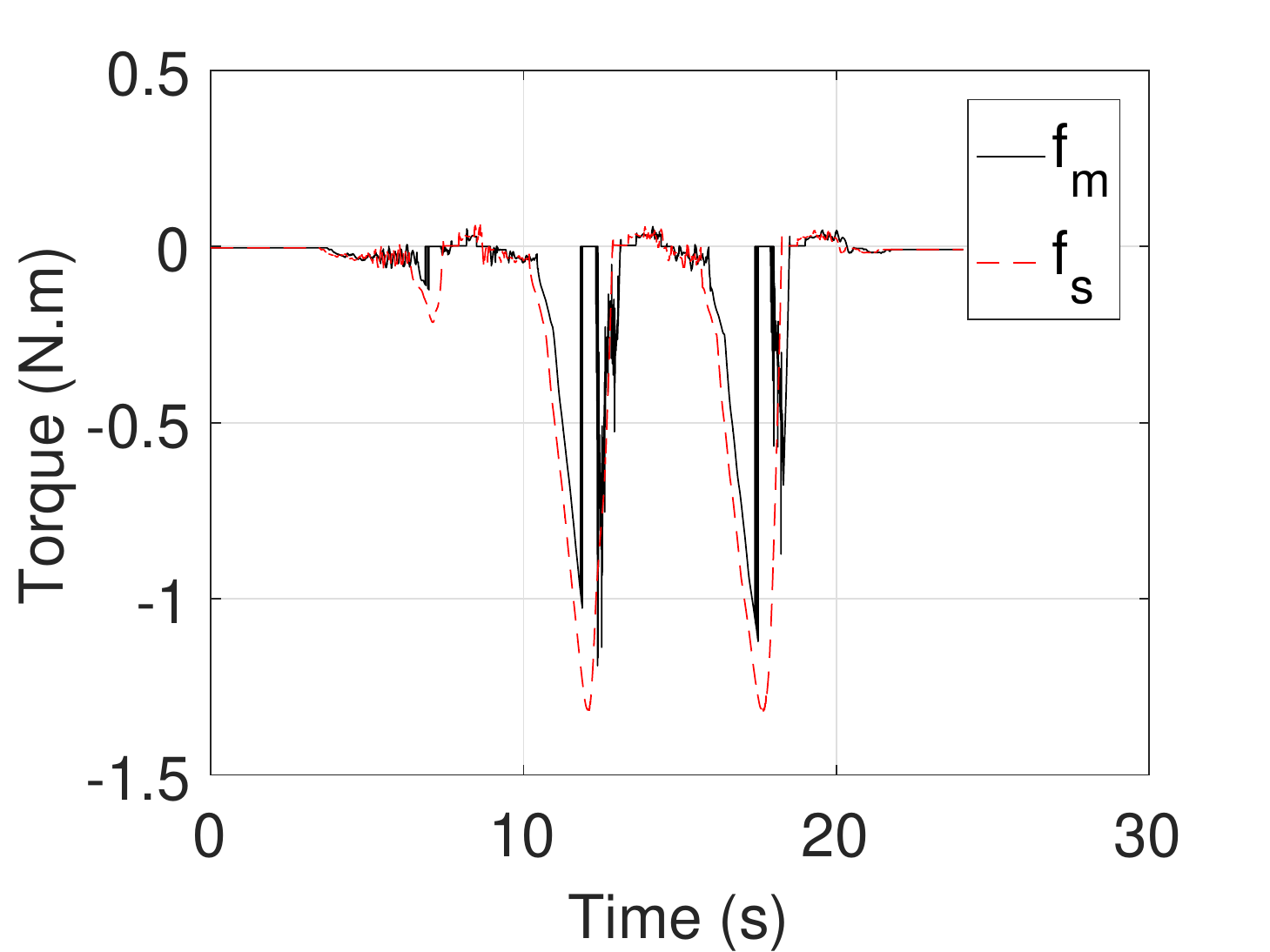}
\caption{}
\label{fig:tdpa500_2}
\end{subfigure}

\begin{subfigure}[thpb]{0.5\linewidth}
\includegraphics[trim={0.2cm 0cm 1.1cm 0.6cm},clip,width=1.5in]{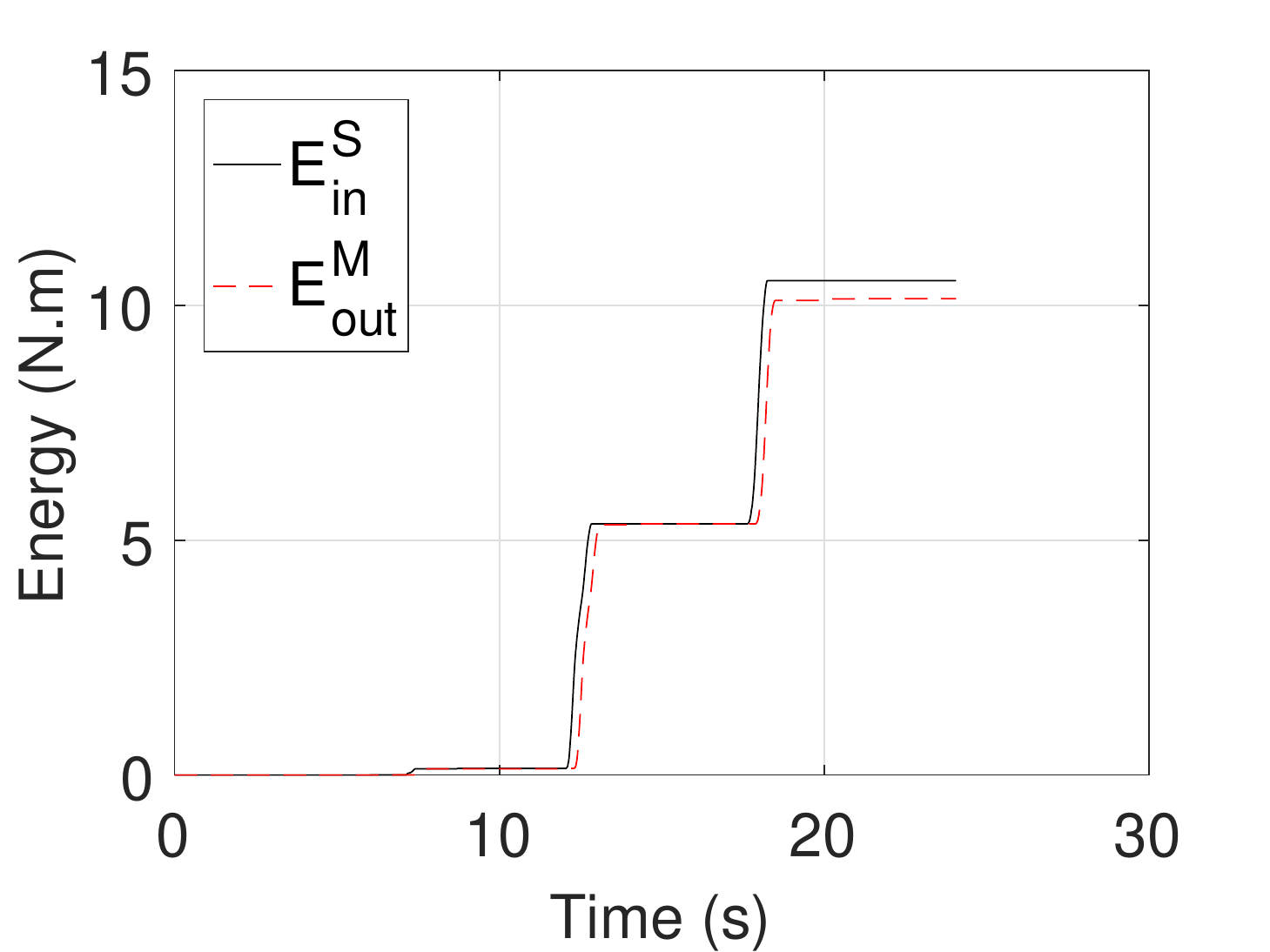}
\caption{}
\label{fig:tdpa500_3}
\end{subfigure}%
~
\begin{subfigure}[thpb]{0.5\linewidth}
\includegraphics[trim={0.2cm 0cm 1.1cm 0.6cm},clip,width=1.5in]{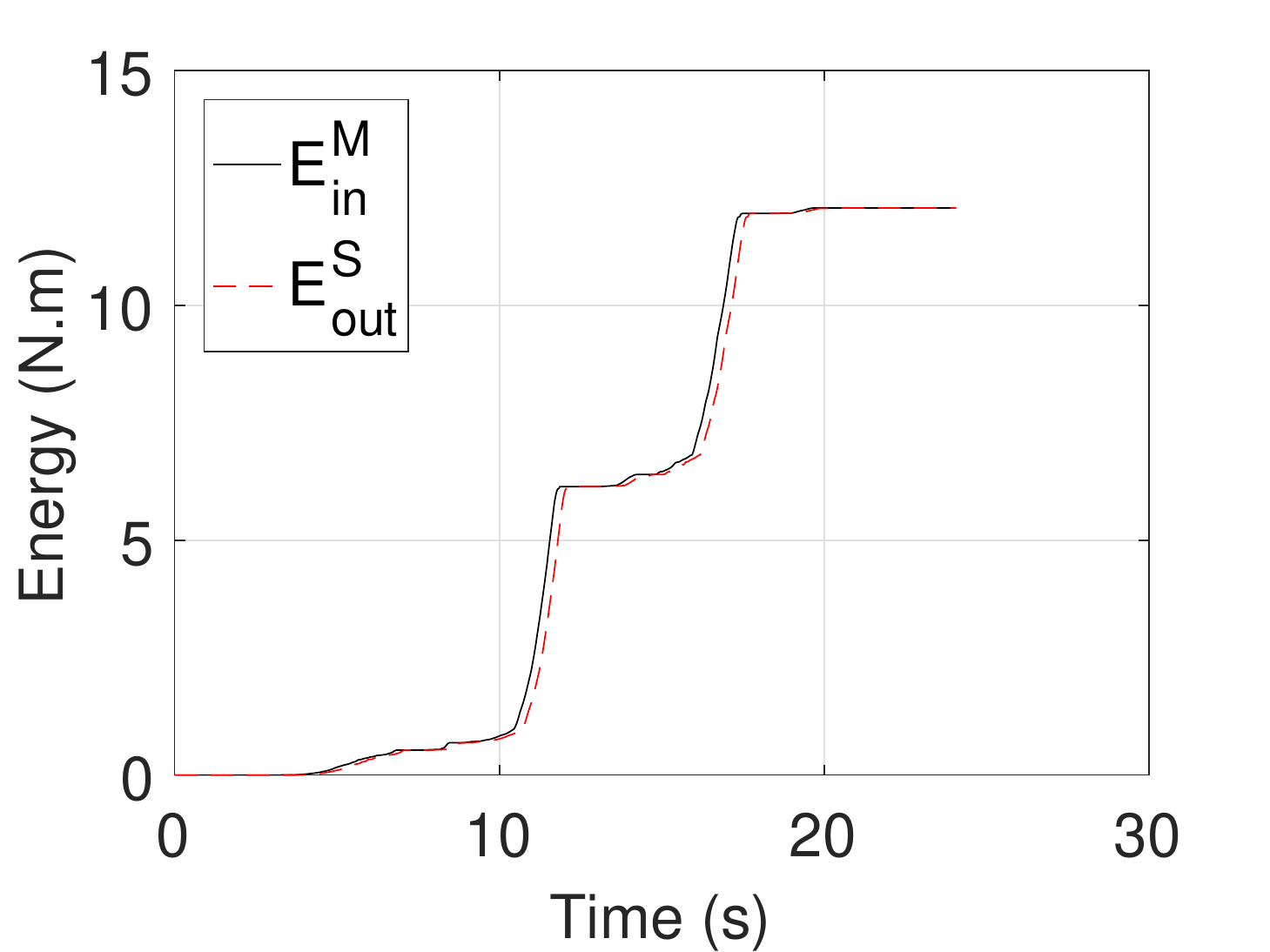}
\caption{}
\label{fig:tdpa500_4}
\end{subfigure}
\caption{$T_{rt}=500$ ms wall contact -- no drift compensator}
\label{fig:tdpa500}
\end{figure}

\begin{figure}[thpb]
\centering
\begin{subfigure}[thpb]{0.5\linewidth}
\includegraphics[trim={0.2cm 0cm 1.1cm 0.6cm},clip,width=1.5in]{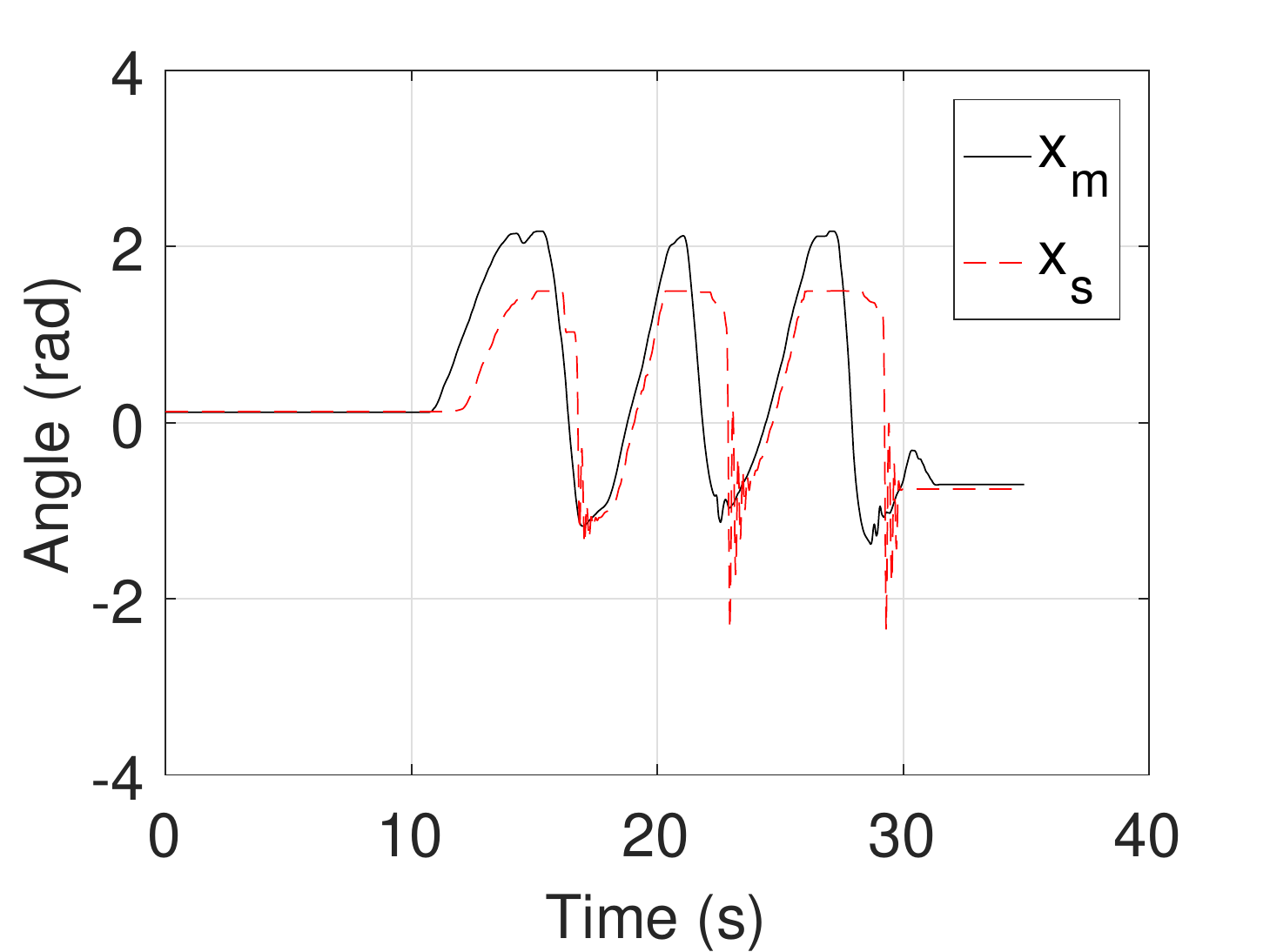}
\caption{}
\label{fig:chawda500_1}
\end{subfigure}%
~
\begin{subfigure}[thpb]{0.5\linewidth}
\includegraphics[trim={0.2cm 0cm 1.1cm 0.6cm},clip,width=1.5in]{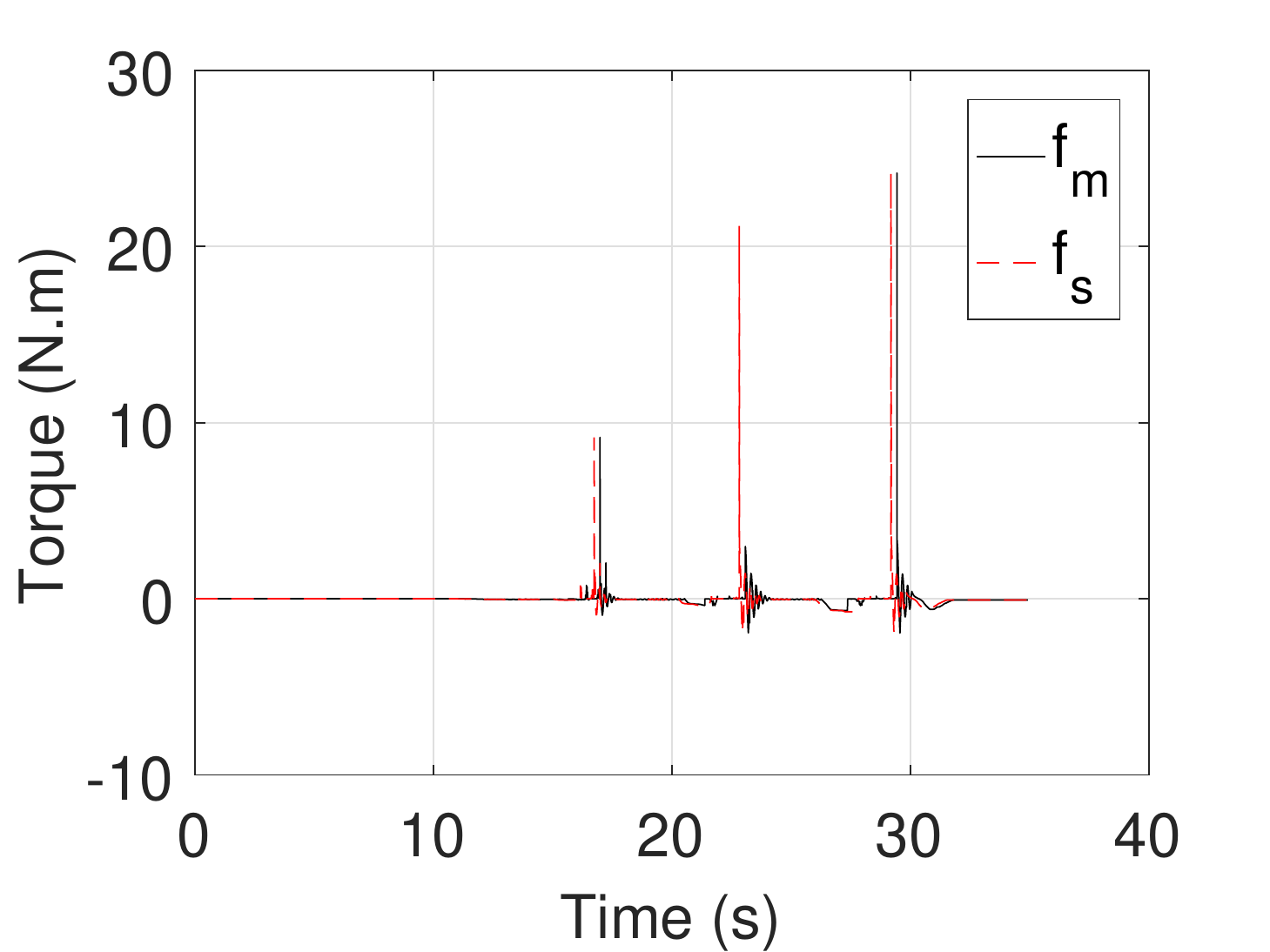}
\caption{}
\label{fig:chawda500_2}
\end{subfigure}

\begin{subfigure}[thpb]{0.5\linewidth}
\includegraphics[trim={0.2cm 0cm 1.1cm 0.6cm},clip,width=1.5in]{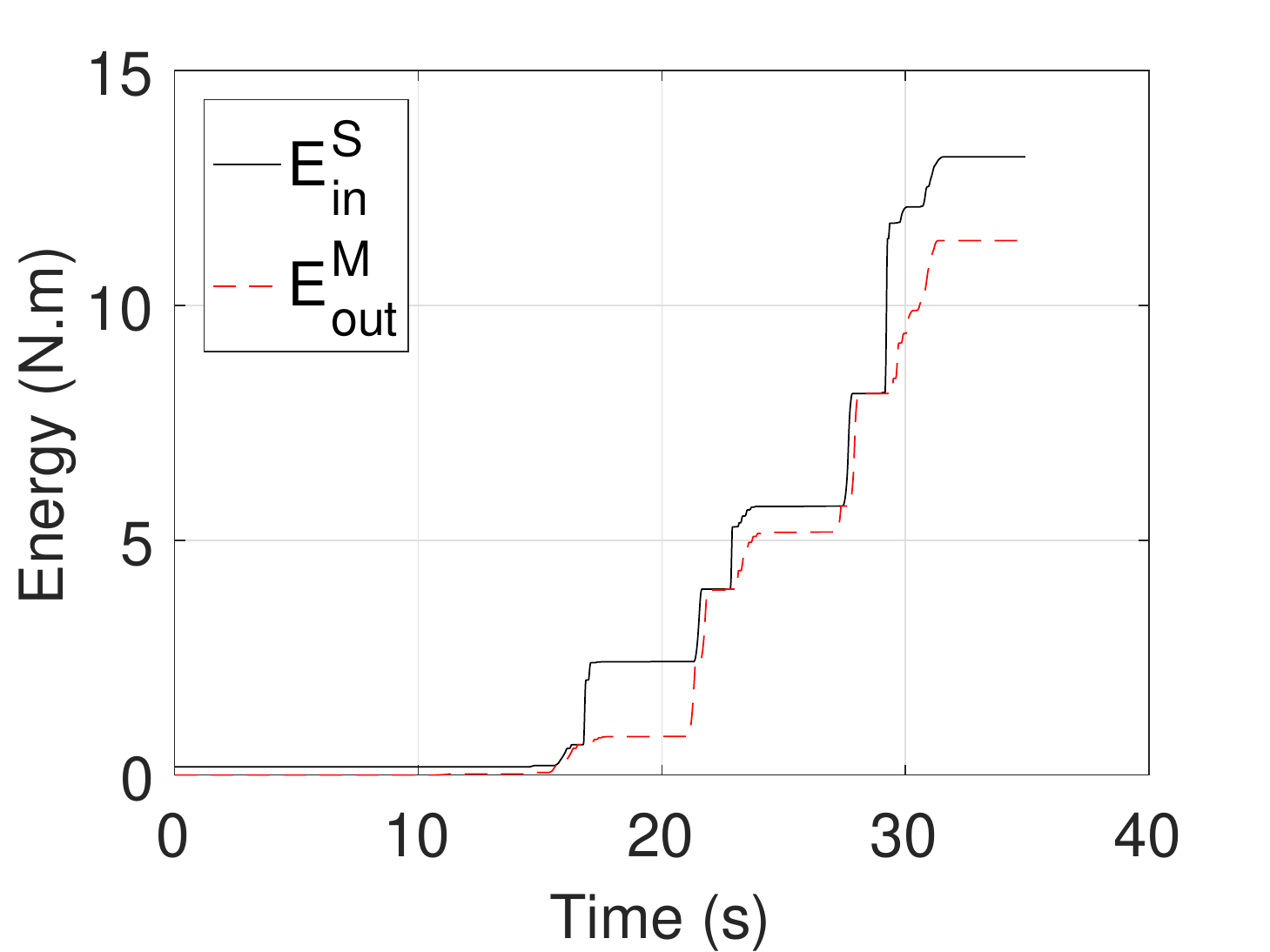}
\caption{}
\label{fig:chawda500_3}
\end{subfigure}%
~
\begin{subfigure}[thpb]{0.5\linewidth}
\includegraphics[trim={0.2cm 0cm 1.1cm 0.6cm},clip,width=1.5in]{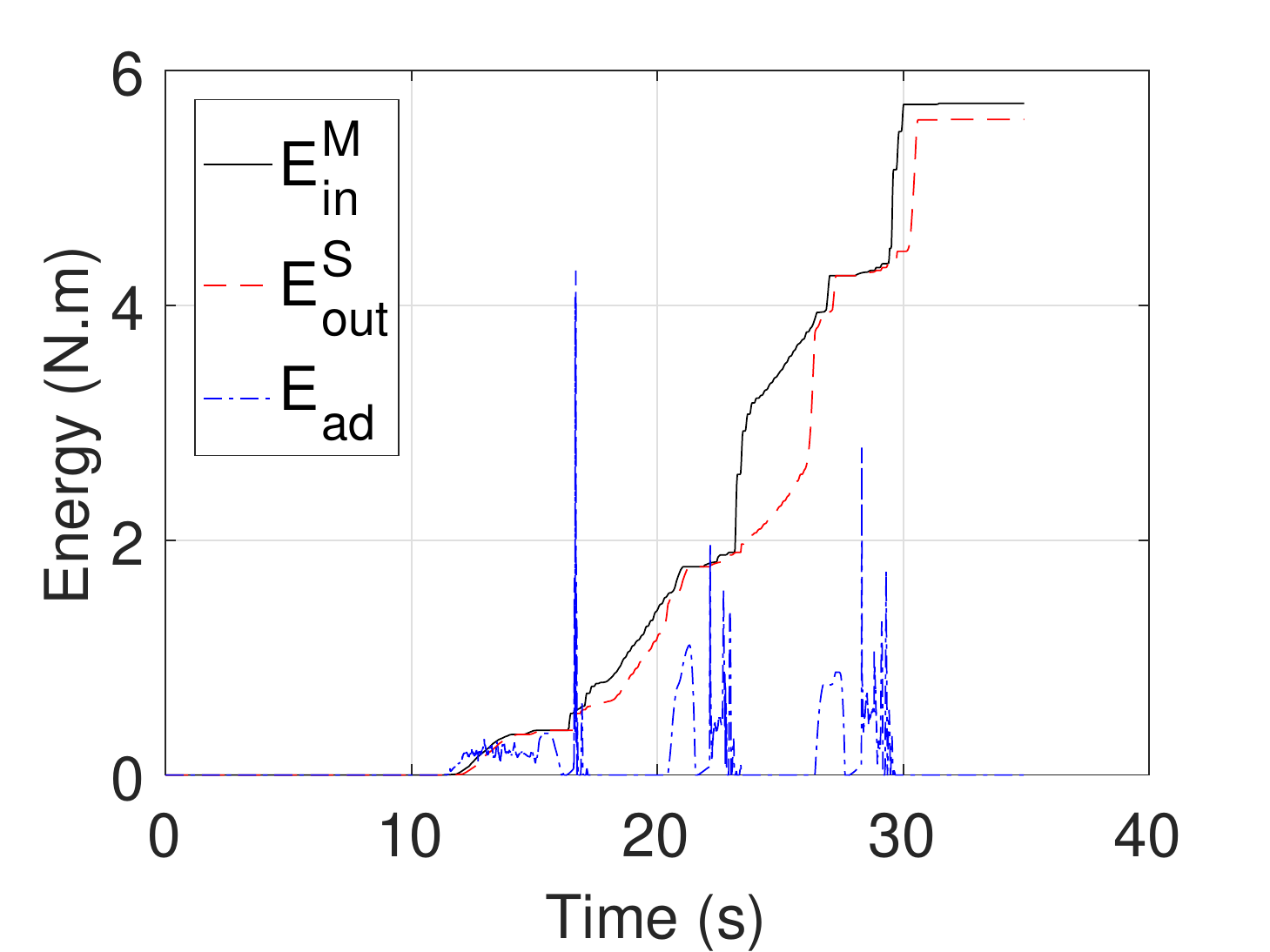}
\caption{}
\label{fig:chawda500_4}
\end{subfigure}
\caption{$T_{rt}=500$ ms wall contact -- compensator from \cite{chawda14}}
\label{fig:chawda500}
\end{figure}

\begin{figure}[thpb]
\centering
\begin{subfigure}[thpb]{0.5\linewidth}
\includegraphics[trim={0.2cm 0cm 1.1cm 0.6cm},clip,width=1.5in]{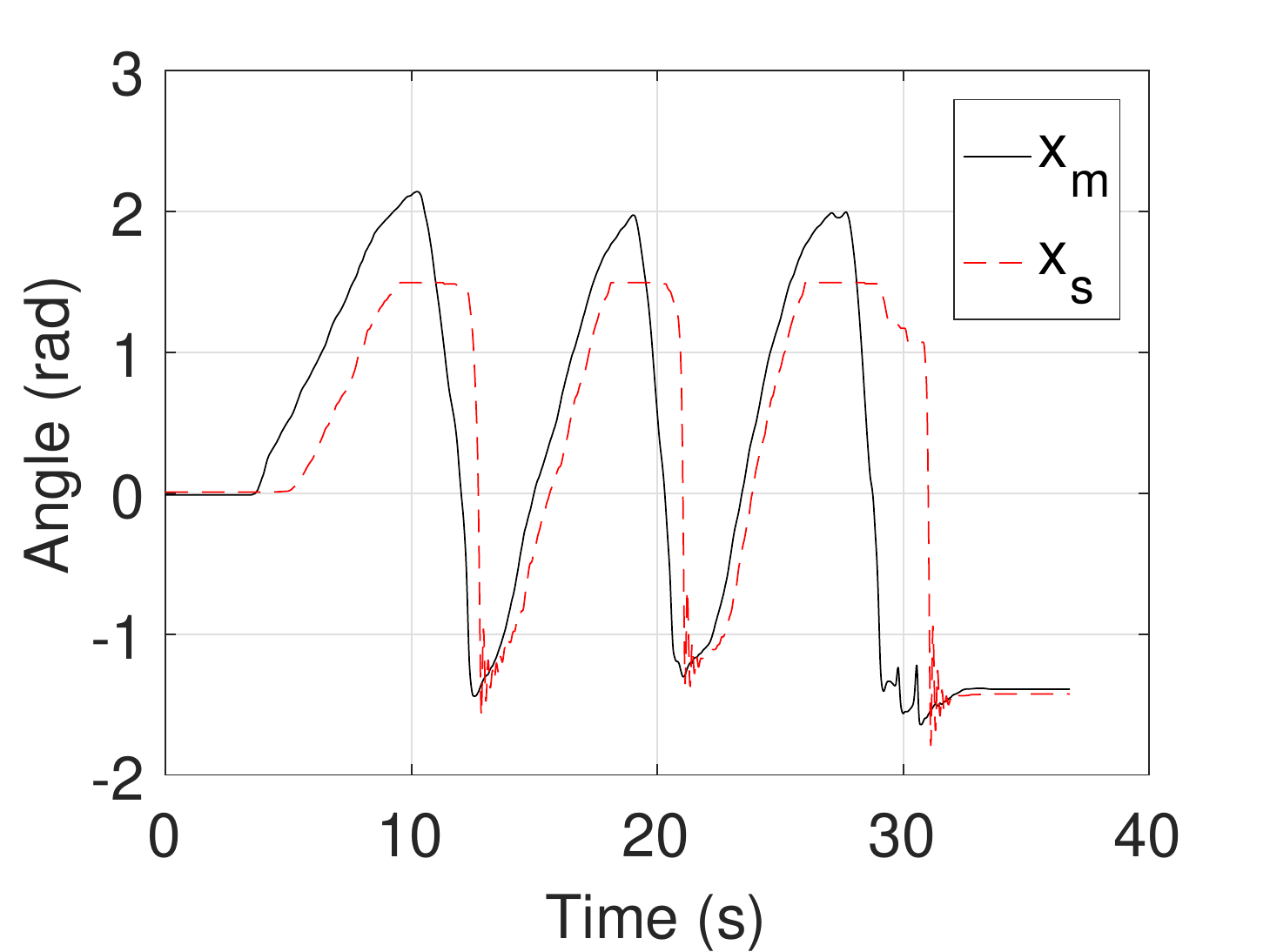}
\caption{}
\label{fig:andre500_1}
\end{subfigure}%
~
\begin{subfigure}[thpb]{0.5\linewidth}
\includegraphics[trim={0.2cm 0cm 1.1cm 0.6cm},clip,width=1.5in]{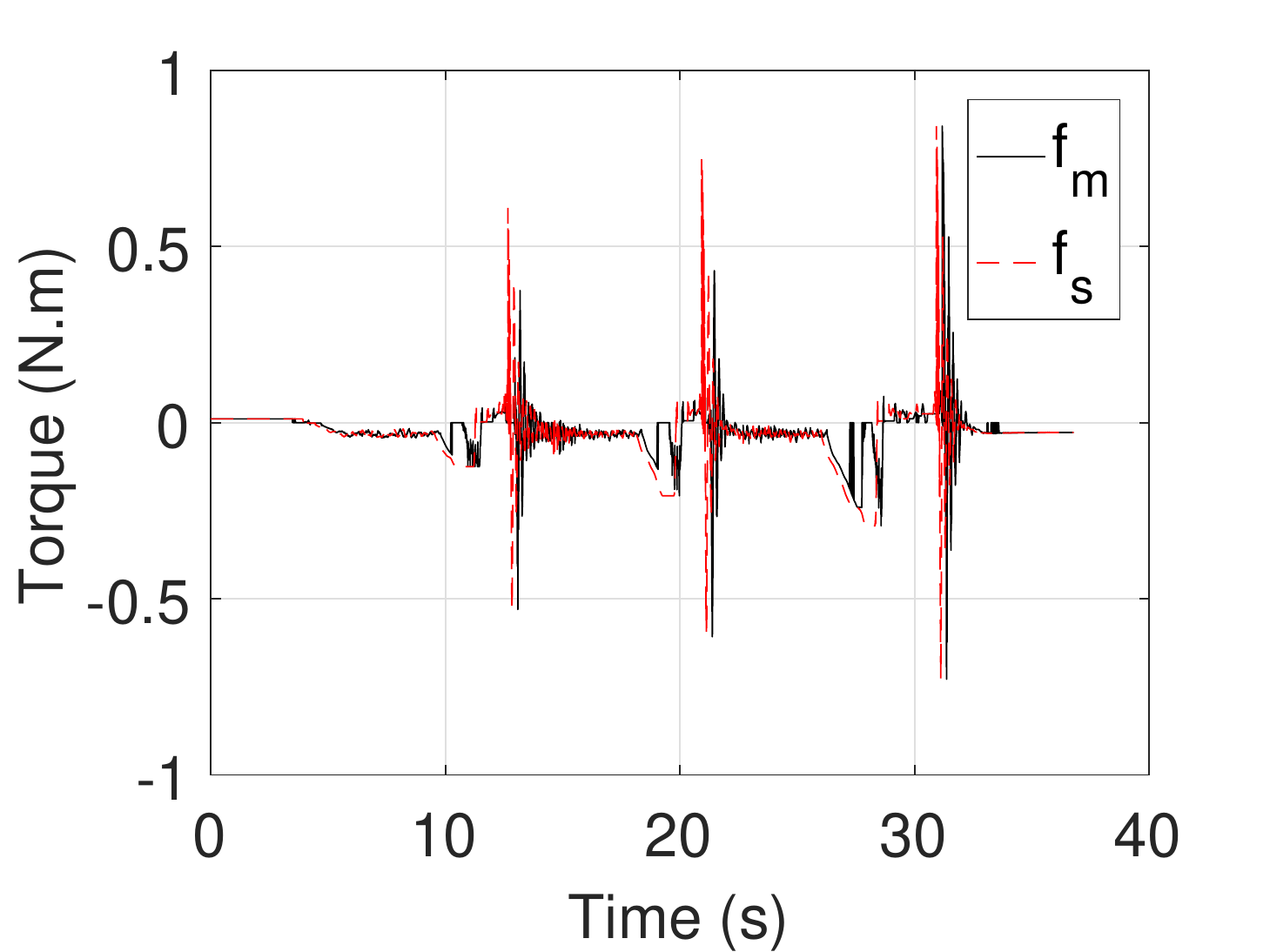}
\caption{}
\label{fig:andre500_2}
\end{subfigure}

\begin{subfigure}[thpb]{0.5\linewidth}
\includegraphics[trim={0.2cm 0cm 1.1cm 0.6cm},clip,width=1.5in]{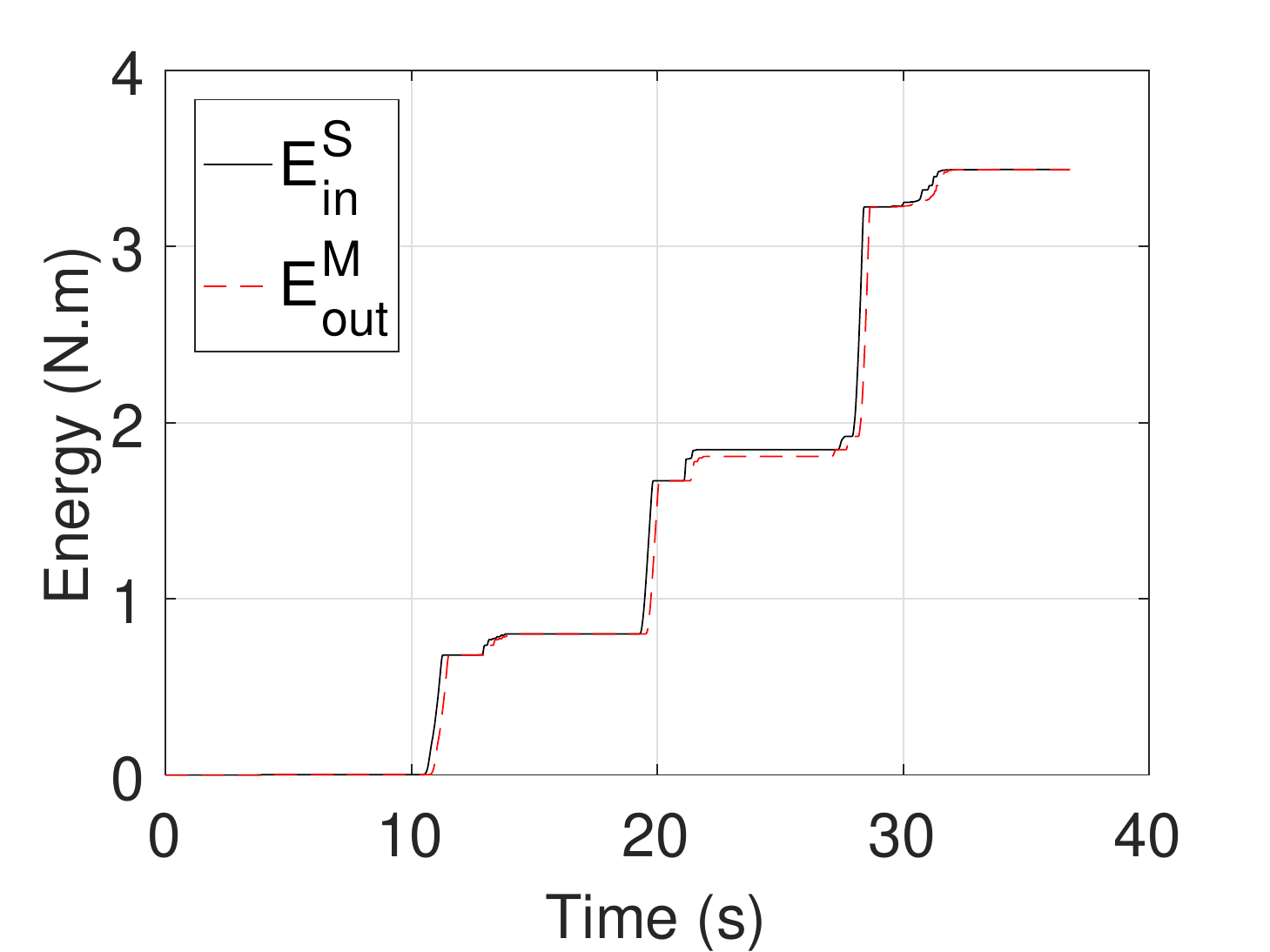}
\caption{}
\label{fig:andre500_3}
\end{subfigure}%
~
\begin{subfigure}[thpb]{0.5\linewidth}
\includegraphics[trim={0.2cm 0cm 1.1cm 0.6cm},clip,width=1.5in]{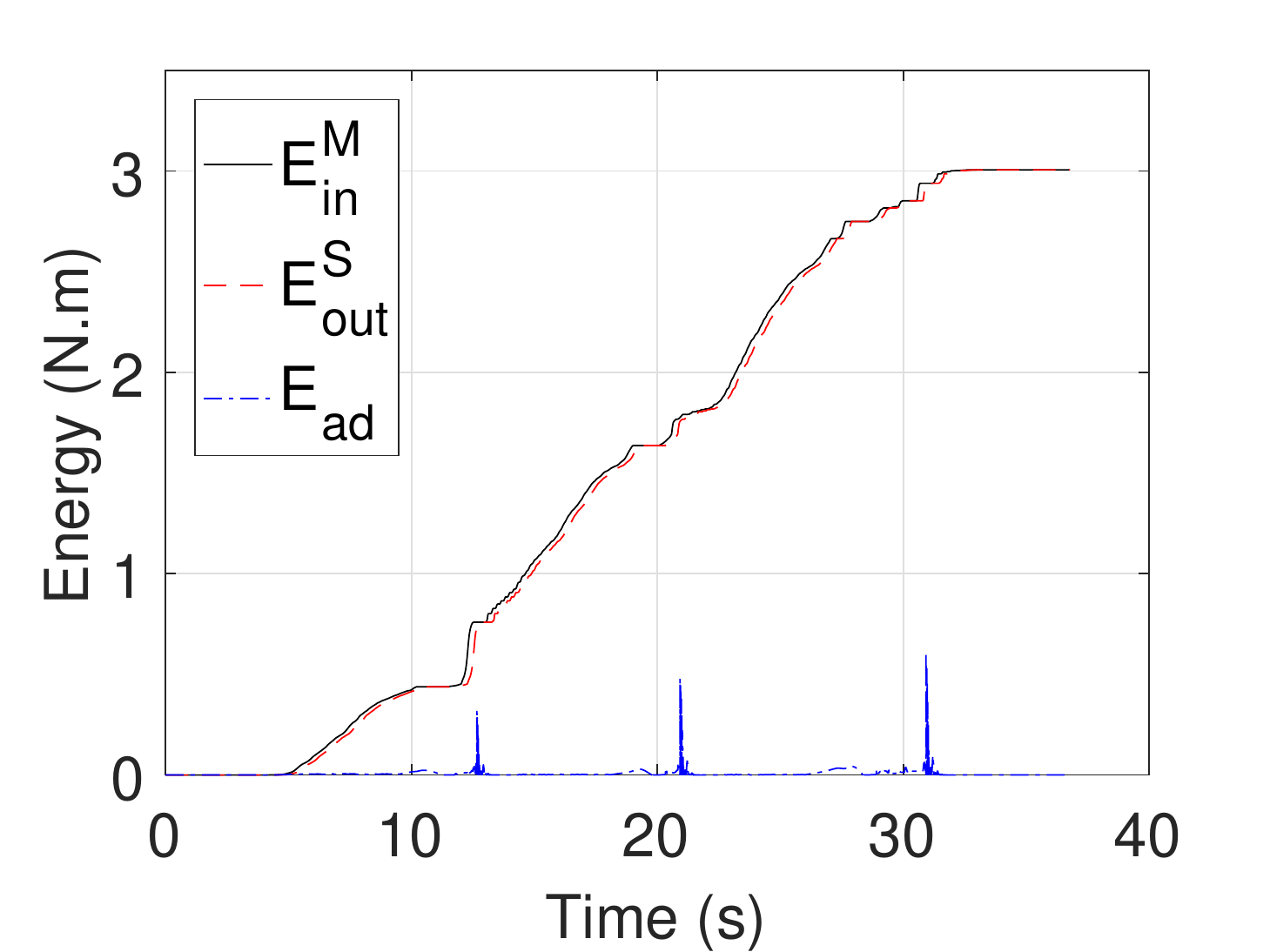}
\caption{}
\label{fig:andre500_4}
\end{subfigure}
\caption{$T_{rt}=500$ ms wall contact -- proposed compensator}
\label{fig:andre500}
\end{figure}

The experiments conducted for $T_{rt}=500$ ms (Figs.~\ref{fig:tdpa500}, \ref{fig:chawda500}, and \ref{fig:andre500}) show that both compensators are able to remove the position drift (Figs.~\ref{fig:chawda500_1} and \ref{fig:andre500_1}). However, for 500 ms the system tends to become very active due to the low inertia and damping of the devices used. To dissipate the extra energy the PC acts at almost all time steps, which gives both compensators few opportunities to act. Because of that, the human operator had to add extra low amplitude movements to the sinusoidal reference in order to create extra passivity gaps (Figs.~\ref{fig:chawda500_1} and \ref{fig:andre500_1}).  Note that the value of $T_{rt}$ where passivity gaps start becoming rare depends on the damping and inertia of the device. A larger or more damped device could allow for teleoperation with higher delays without the operator having to care about passivity gaps. Comparing the behavior of both compensators, it can be noted that, even though some torque spikes also start to become present in the proposed approach due to drift accumulation (Fig.~\ref{fig:andre500_2}), those are still close to the range of the normal teleoperation torques. On the other hand, the compensator from \cite{chawda14} presents higher torques (Fig.~\ref{fig:chawda500_2}), which are much more perceptible and could be even dangerous to the human operator.  

The fact that force spikes generated by the method described in \cite{chawda14} are much lower in the proposed approach is due to two reasons. First, the proposed compensator checks the \textit{untouched} delayed velocity $\hat{v}_{sd}$ from the master and the current velocity of the slave $v_s$, which makes the compensator correct only the actual drift observed between the devices. Having the compensator guarantee that the slave controller gets the exact information about the position of the master is also able to solve the position drift problem. However, unnecessary compensation action could also be applied if the dynamics of the task would make the drift smaller than the difference between the \textit{untouched} and the reference position. Since the PC acts as a variable saturation to the \textit{untouched} velocity, the signal of $x_{err}$ from (\ref{eq:x_err}) would exhibit abrupt jumps, while the correction from the proposed compensator (\ref{eq:comp}) would be smoother since the controller and the slave device would filter the high frequency dynamics. Moreover, adding a gain $K$ to the compensator equation makes this approach more versatile to different tasks since the gain could be increased or decreased according to the nature of the task in order to obtain faster or smoother corrections.

\subsection{Variable Time Delays} 
\begin{figure}[thpb]
\centering
\begin{subfigure}[thpb]{0.5\linewidth}
\includegraphics[trim={0.2cm 0cm 1.1cm 0.6cm},clip,width=1.5in]{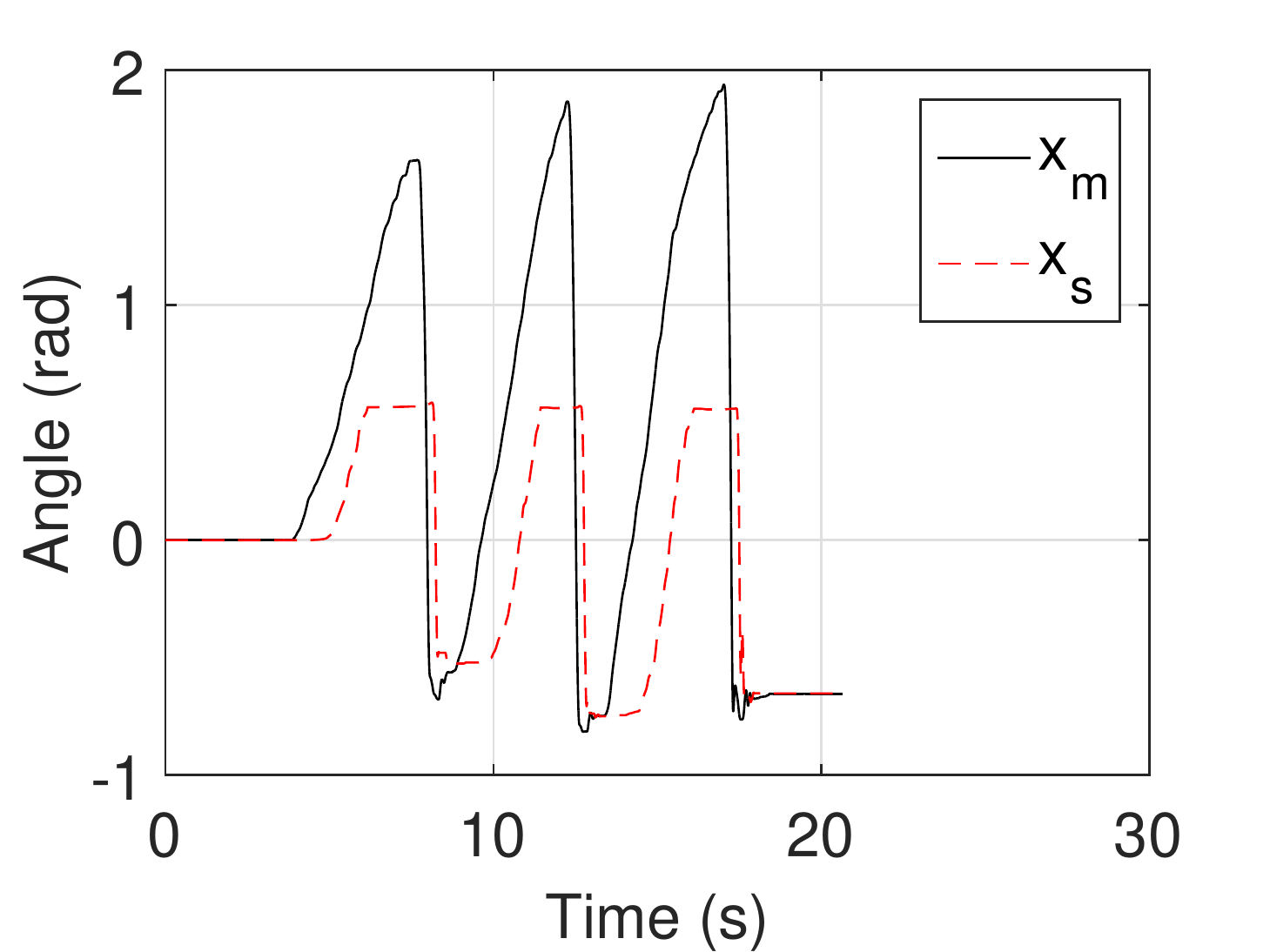}
\caption{}
\label{fig:andre200_1var}
\end{subfigure}%
~
\begin{subfigure}[thpb]{0.5\linewidth}
\includegraphics[trim={0.2cm 0cm 1.1cm 0.6cm},clip,width=1.5in]{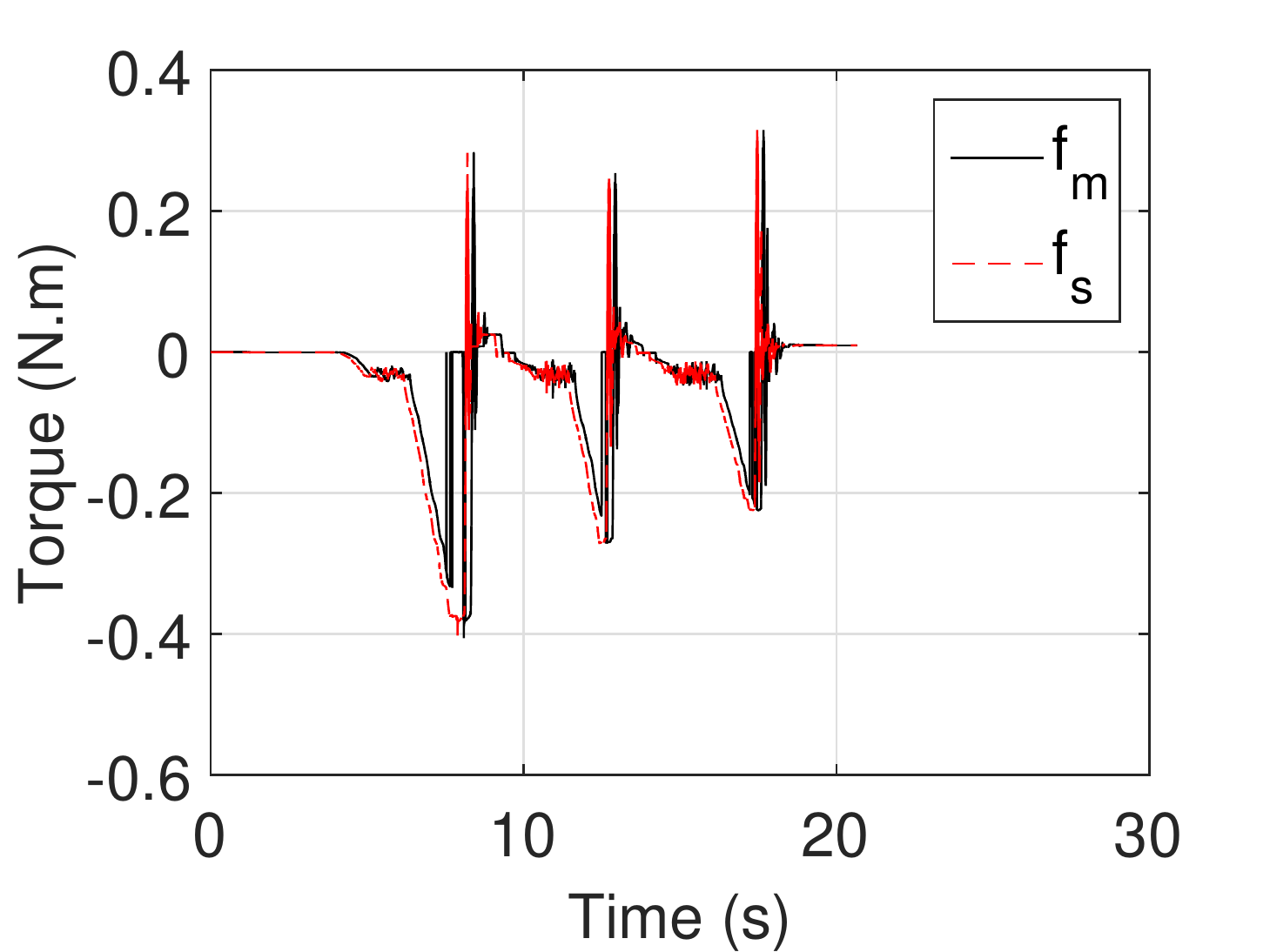}
\caption{}
\label{fig:andre200_2var}
\end{subfigure}

\begin{subfigure}[thpb]{0.5\linewidth}
\includegraphics[trim={0.2cm 0cm 1.1cm 0.6cm},clip,width=1.5in]{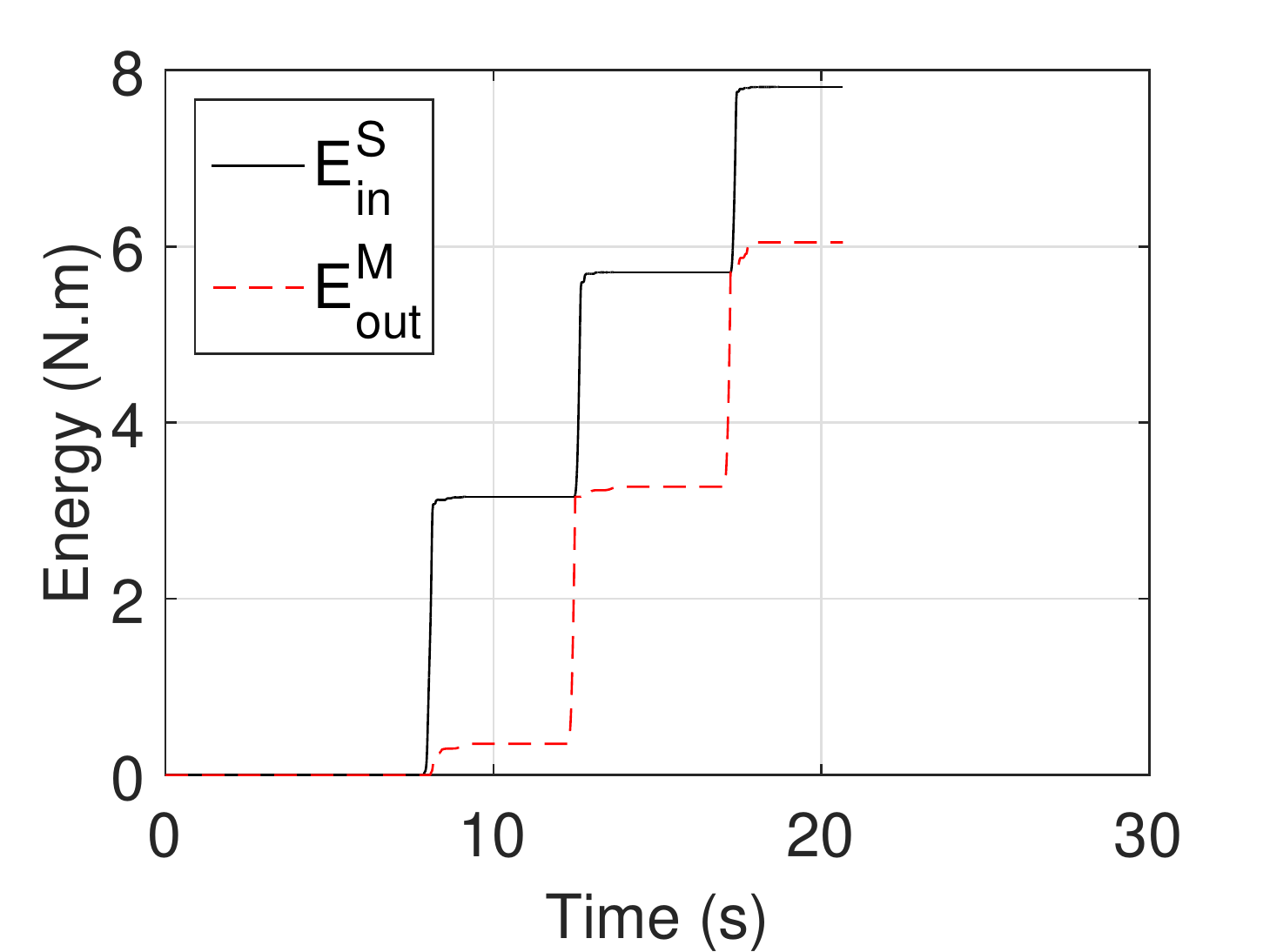}
\caption{}
\label{fig:andre200_3var}
\end{subfigure}%
~
\begin{subfigure}[thpb]{0.5\linewidth}
\includegraphics[trim={0.2cm 0cm 1.1cm 0.6cm},clip,width=1.5in]{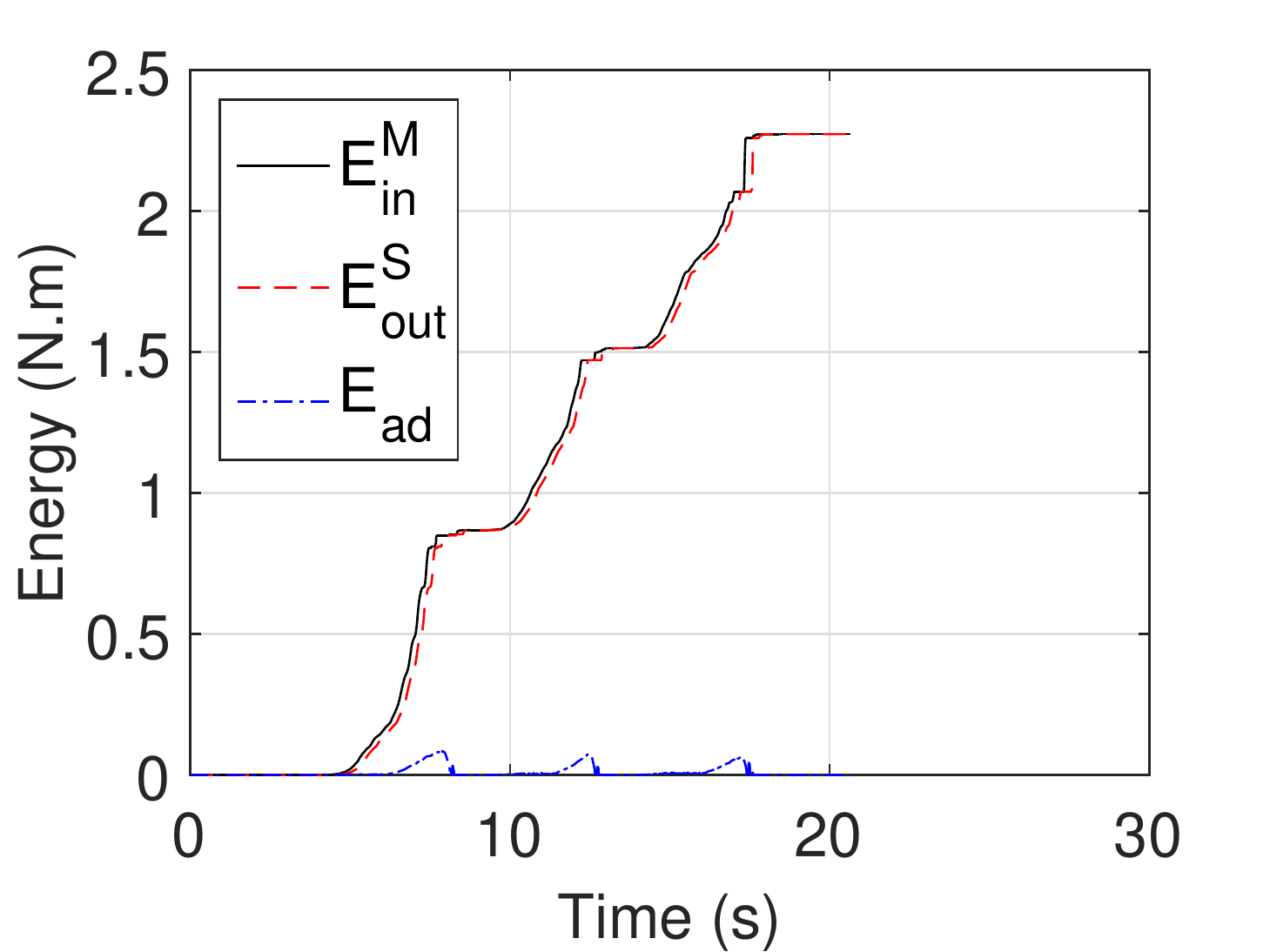}
\caption{}
\label{fig:andre200_4var}
\end{subfigure}
\caption{$T_{rt}=200$ ms $\pm 10$\% wall contact -- proposed compensator}
\label{fig:andre200var}
\end{figure}

In order to demonstrate the applicability of the proposed compensator to variable delay scenarios, an experiment was conducted for round-trip delays of 200 ms $\pm 10$\%.  Fig.~\ref{fig:andre200var} displays the results of the experiment. It can be noted that, despite the variable delays, the compensator was able to eliminate the position drift (Fig.~\ref{fig:andre200_1var}), while keeping the forces at their normal range (Fig.~\ref{fig:andre200_2var}). It can be also seen that the energy added by the compensator ($E_{ad}$) is small compared to the energy flow in the channel. Since this compensation method uses energy-based TDPA, it is able to deal with large delay variations while keeping the system stable (see Section~\ref{sec:energy_based}).

\section{CONCLUSIONS}
Although some other approaches (see \cite{chawda14,artigas10}) have been developed in order to eliminate the well-known position drift problem in energy-based TDPA, those exhibited force spikes that affect the natural feeling of teleoperation and can be harmful to the operator and the hardware.This paper proposes a novel TDPA-based drift compensator, which is able to correct the position drift caused by the passivity controller without generating high forces or torques. Experimental results have proven the feasibility and performance of the proposed approach for round-trip delays of 200 ms, and 500 ms. A comparison with the previous approaches has shown much smaller added forces for the proposed compensator, which provides a smoother operational feel while being able to correct the position drift in bilateral teleoperation. \par
The proposed approach carries along the advantages of robustness and adaptiveness of energy-based TDPA and is also able to solve a drawback of this formulation, namely position drift, without generating high impulse-like forces.The compensator can also be tuned to be used in different tasks, where fast drift compensation or low forces are necessary. The limitations of this method arise from the nature of TDPA. In order to conserve system passivity, the compensator has to wait until a passivity gap shows up to perform its action, which for very light or very high delays may become a problem.
\par
The use of the proposed approach is not limited to classical teleoperation applications. A similar approach has also been applied to the cooperative landing scenario described in \cite{muskardin17}. Further details on this application will be presented in the near future. 
\par
Future work will also involve applying the proposed approach to multi-DoF devices and presenting an adaptive law for tuning the compensator.

\end{document}